\documentclass[acmtog]{acmart}

\usepackage{booktabs} 

\usepackage[ruled]{algorithm2e} 

\SetAlFnt{\small}
\SetAlCapFnt{\small}
\SetAlCapNameFnt{\small}
\SetAlCapHSkip{0pt}

\setcopyright{rightsretained} 
\acmJournal{TOG}
\acmYear{2021}
\acmVolume{40}
\acmNumber{4}
\acmArticle{143}
\acmMonth{8} 
\acmDOI{10.1145/3450626.3459850}


\usepackage{bm}
\usepackage{comment}

\usepackage{graphicx}

\begin{document}

\title{Driving-Signal Aware Full-Body Avatars}


\author{Timur Bagautdinov}
\affiliation{
\institution{Facebook Reality Labs}
\city{Pittsburgh}
\country{USA}
}
\email{timurb@fb.com}

\author{Chenglei Wu}
\affiliation{
\institution{Facebook Reality Labs}
\city{Pittsburgh}
\country{USA}
}
\email{chenglei@fb.com}

\author{Tomas Simon}
\affiliation{
\institution{Facebook Reality Labs}
\city{Pittsburgh}
\country{USA}
}
\email{tsimon@fb.com}

\author{Fabián Prada}
\affiliation{
\institution{Facebook Reality Labs}
\country{USA}
}
\email{fabianprada@fb.com}

\author{Takaaki Shiratori}
\affiliation{
\institution{Facebook Reality Labs}
\city{Pittsburgh}
\country{USA}
}
\email{tshiratori@fb.com}

\author{Shih-En Wei}
\affiliation{
\institution{Facebook Reality Labs}
\city{Pittsburgh}
\country{USA}
}
\email{swei@fb.com}

\author{Weipeng Xu}
\affiliation{
\institution{Facebook Reality Labs}
\city{Pittsburgh}
\country{USA}
}
\email{xuweipeng@fb.com}

\author{Yaser Sheikh}
\affiliation{
\institution{Facebook Reality Labs}
\city{Pittsburgh}
\country{USA}
}
\email{yasers@fb.com}

\author{Jason Saragih}
\affiliation{
\institution{Facebook Reality Labs}
\country{USA}
}
\email{jsaragih@fb.com}



\begin{teaserfigure}
    \centering
    \includegraphics[width=\linewidth]{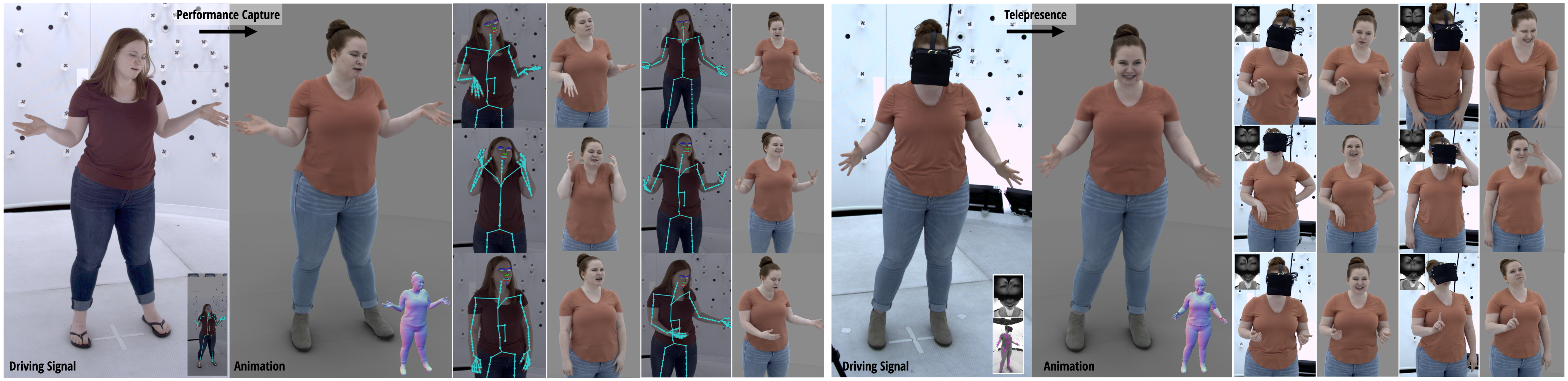}
    \vspace{-0.6cm}
    \caption{
    We present an approach for building photorealistic full-body avatars that can be animated using input driving signals such as 3D keypoints (left) or body pose and facial animation codes (right). We build a disentangled latent space that is \emph{driving-signal aware} to ensure that the model generalizes to novel sequences while still producing photorealistic results. 
    }
    \label{fig:teaser}
\end{teaserfigure}

\begin{abstract}
We present a learning-based method for building driving-signal aware full-body avatars.
Our model is a conditional variational autoencoder that can be animated with incomplete 
driving signals, such as human pose and facial keypoints, and produces 
a high-quality representation of human geometry and view-dependent appearance.
The core intuition behind our method is that better drivability and generalization
can be achieved by disentangling the driving signals and remaining generative factors,
which are not available during animation.
To this end, we explicitly account for information deficiency in the driving signal 
by introducing a latent space that exclusively captures the remaining information,
thus enabling the imputation of the missing factors required during full-body animation, 
while remaining faithful to the driving signal.
We also propose a learnable localized compression for the driving signal which
promotes better generalization, and helps minimize the influence 
of global chance-correlations often found in real datasets. 
For a given driving signal, the resulting variational model produces a compact space of uncertainty 
for missing factors that allows for an imputation strategy best suited to a particular application. 
We demonstrate the efficacy of our approach on the challenging problem of full-body animation for 
virtual telepresence with driving signals acquired from minimal sensors placed in the environment
and mounted on a VR-headset. 
\end{abstract}
%
%

\newcommand{\JS}[1]{{\color{magenta}{\bf JS: #1}}}
\newcommand{\TB}[1]{{\color{green}{\bf TB: #1}}}
\newcommand{\tb}[1]{{\color{green} #1}}
\newcommand{\CW}[1]{{\color{blue}{\bf CW: #1}}}
\newcommand{\cw}[1]{{\color{blue} #1}}

\newcommand{\updated}[1]{{\color{red} #1}}


\newcommand{\bb}{\mathbf{b}}
\newcommand{\bx}{\mathbf{x}}
\newcommand{\by}{\mathbf{y}}
\newcommand{\bz}{\mathbf{z}}
\newcommand{\bc}{\mathbf{c}}
\newcommand{\bbf}{\mathbf{f}}
\newcommand{\bv}{\mathbf{v}}
\newcommand{\be}{\mathbf{e}}
\newcommand{\bt}{\mathbf{t}}

\newcommand{\bT}{\mathbf{T}}
\newcommand{\bG}{\mathbf{G}}
\newcommand{\bW}{\mathbf{W}}
\newcommand{\bR}{\mathbf{R}}
\newcommand{\bM}{\mathbf{M}}
\newcommand{\bI}{\mathbf{I}}

\newcommand{\balpha}{\bm{\alpha}}
\newcommand{\bbeta}{\bm{\beta}}
\newcommand{\bgamma}{\bm{\gamma}}
\newcommand{\bdelta}{\bm{\delta}}
\newcommand{\btheta}{\bm{\theta}}

\newcommand{\bmu}{\bm{\mu}}
\newcommand{\bsigma}{\bm{\sigma}}

\newcommand{\mR}{\mathbb{R}}
\newcommand{\mL}{\mathcal{L}}
\newcommand{\mN}{\mathcal{N}}

\newcommand{\ttt}[1]{\texttt{#1}}


\maketitle

\section{Introduction}  

The goal of this work is to build high quality full body models of geometry and appearance that can be driven from commodity sensors placed in the environment.
Building expressive and animatable virtual humans is a well studied problem in the graphics community. The creation of so called digital doubles has roots in the special effects industry~\cite{Emily10}, and in recent years has begun to see examples of real-time uses as well such as Siren of Epic Games and DigiDoug of Digital Domain. These models are typically built using sophisticated multi-view capture systems with elaborate scripts that span variations in pose and expression. 
They capture a low-dimensional prior representation of human shape and appearance that allows for animation using more modest capture settings~\cite{Blanz99,loper2015smpl,MANO2017}. 
Nonetheless, these models rely on brittle hand-crafted assumptions to encourage generalization to unseen poses, and overall sensor requirements for animation remains significant (e.g., a full-body motion capture suit for animation such as Xsens\footnote{https://www.xsens.com/} and OptiTrack\footnote{https://www.optitrack.com/}). Since these models are learned independently from the sensor configurations used to drive the model during animation, they can simultaneously be over- and under-constrained, with limited ability to precisely match poses observed by the sensors while exhibiting unrealistic contortions in unobserved areas. 

One approach to better integrate the driving signal into the model building process is to simultaneously capture using both the modeling sensors 
(i.e. an elaborate multi-view capture system) and the animation sensors (i.e. a small set of cameras placed in the environment). 
In this scenario, one can learn a model that directly regresses full shape and appearance from
the information-deficient driving signals captured by the animation sensors. 
Although this approach significantly restricts the types of driving signals that can be used,
e.g. a motion capture suit for a regularly clothed digital double, we argue that there is a more fundamental 
problem with this scenario.
That is, there exists information asymmetry between the modeling and animation sensors which 
results in a one-to-many mapping problem, where multiple combinations of model states can equally 
likely explain the measurements. 
For example, a driving signal based on body joint angles does not contain complete information about 
clothing wrinkles and muscle contraction. Similarly, facial keypoints typically do not encode the hair, gaze or tongue motion. 
As a result, a model that is trained naively, without specifically taking into account
such missing information, will either under-fit and produce averaged appearance, or, 
given enough capacity, will over-fit and learn chance correlations only present in 
the training set. 
Some existing works have addressed the issue of information asymmetry through the use of temporal models and adversarial training~\cite{ginosar2019gestures,alexanderson2020speech2gesture,ng2020body2hands,yoon_2020_TOG}. 
However, these methods tend to prescribe a specific imputation strategy that may not be appropriate in some applications. 
Furthermore, they operate on models post-training, making it difficult to overcome over- or under-fitting behavior 
that may already exist in the model. 

Our work aims to address the problem of learning a model for a full-body digital avatar that is faithful to
information-deficient driving signals while also providing an explicit data-driven space of plausible configurations
for the missing information.
To this end, we introduce a variational model that explicitly captures two types of factors of variation: 
\textit{observed} factors that can be reliably estimated from the driving signals during animation, 
and \textit{missing} factors, which are available only during the modeling stage. 
Our core strategy is to encourage better generalization by minimizing the correlations between the observed 
factors, while maximizing it for the missing factors, such that during animation the model is able to produce
plausible/realistic appearance and shape configurations that are fully consistent with the driving signal. 
We achieve the first by building a spatially-varying representation of the driving signal
that localizes its effects and breaks global chance-correlations that might exist in the training set. 
The second is achieved by introducing a latent space for variation that is disentangled from the observed factors, forcing it to capture only the missing factors that are necessary to reconstruct the data.
In particular, better disentanglement is achieved by explicitly accounting for coarse-long range effects of the driving signal that would otherwise be modeled by the latent space due to the localized conditioning used to encourage generalization. We achieve this using a coarse model of limb motion and an ambient occlusion map that helps model self-shadowing without overfitting.
The resulting model can generate the space of plausible animations that agree with the information contained in the driving signal (Figure~\ref{fig:teaser}).
Because of the explicit separation between the observed and missing factors, our approach enables
the freedom to employ imputation techniques best suited to a particular application. 
We demonstrate the effectiveness a particularly simple approach; we assign the mean value to the missing factors for all frames during a sequence, which results in compelling animations that avoids over- or under-fitting effects observed in other approaches.

To summarize, the contributions of this work are as follows:
\begin{itemize}
    \item A representation for a full body model that explicitly accounts for the driving signal during its construction. The model can generate a diverse space of plausible configurations that agree with the information contained in the driving signal.
    \item A method for achieving good generalization to novel inputs while producing high quality reconstructions by employing localized conditioning, accounting for coarse long range effects, and disentangling driving signals from the latent space.
    \item A demonstration of the utility of this approach on two scenarios where driving signal information is deficient: performance capture with a different attire and avatar animation for VR telepresence.
\end{itemize}

\section{Related Work}

Our main goal is to build personalized full-body avatars that can
be animated with information-deficient driving signals, while providing flexibility
to practitioners to impute missing information appropriately
for the application at hand.

\subsection{Avatar Modeling}

In the last decade, many efforts have been made for achieving expressive and animatable 3D models for human avatars, including face, hands and body. Due to the complexity in geometric deformation and appearance, data-driven methods have become popular~\cite{Blanz99,loper2015smpl,MANO2017}. 
Linear models, e.g. PCA or Blendshapes, have been employed to model the muscle-activated skin deformation space, and were demonstrated to be effective for facial models~\cite{Blanz99,Lau09,Vlasic05,Lewis2014}. 
For hands and body, an articulated prior is usually explicitly modeled by a kinematic chain, 
while the surface deformation associated with the pose is obtained through skinning~\cite{Lewis2000}, which 
generates the deformed surface by a weighted set of the influences from neighboring joints. 
Combining with the statistical modeling tools, a more expressive model can be developed by learning the pose-dependent 
corrective deformation across different identities for a body~\cite{SCAPE05,loper2015smpl} and a hand ~\cite{MANO2017,moon2020deephandmesh}.
Expressiveness can be further improved by localizing the deformation space~\cite{Tena11, Wu16, STAR:2020}. 
These geometric models enable to learn a shading-free appearance model across different identities a hand~\cite{HTML_eccv2020},
to automatically create a textured full-body avatar from a video~\cite{alldieck2018video,alldieck_3dv2018}, and to render avatars under new poses and camera views by using neural rendering~\cite{SMPLpix:WACV:2020}.
Building on the above, unified avatar models, which model face, hand and body altogether, have been developed, 
including the Frank model~\cite{Han18} and the SMPL-X model~\cite{SMPL-X:2019}. However, even given the limited 
training data, the generated results by these models are still 
underfit and constrained by the fidelity they can produce.

With the advent of deep neural networks, deep generative models have been successfully applied to 
modeling human bodies. 
For example, various convolutional mesh autoencoders have been proposed for building models for 
faces~\cite{COMA2018}, or hands and bodies~\cite{YI2020}. 
Compositional VAEs~\cite{Bagautdinov2018} model facial geometry with a hierarchical deep 
generative model, leading to a more expressive learned space of deformations and better modeling 
of high-frequency detail.
Deep Appearance Models~\cite{lombardi2018deep} learns a personalized variational model for both geometry 
and texture of a face, enabling a photorealistic rendering. 
However, extending those methods to full-body avatars is non-trivial, and, since those do not explicitly
account for the missing information, they are prone to artifacts in scenarios with deficient 
driving signals.

Modeling clothing is another aspect closely related to our work, as we are primarily interested in animating 
clothed bodies. 
%
\cite{Stoll2010video} combine the cloth
simulation with a body model from a multi-view capture so that the clothed body can be animated with the simulation. 
Recently~\cite{Han18} has extended the Frank model to Adam for modeling clothed surface with deformation spaces. 
Simulated dynamic clothes are used to model the interaction between cloth and the underneath body by explicitly factoring 
out the dynamic deformation~\cite{DRAPE2012}.
CAPE~\cite{Qianli20} employs Graph-CNN to dress 3D meshes of human body from SMPL. 
However, it is learnt purely on geometry and only on the scans for sparse sampling of poses, 
and thus cannot be driven to generate photorealistic renderings with dynamic clothing.

Another line of work on animating photorealistic human rendering skips the complicated 3D geometry modeling, 
and rather focuses on synthesizing photorealistic images or videos by solving an image translation problem, e.g., learning the mapping function from joint heatmaps~\cite{aberman2019deep}, 
rendered skeleton~\cite{chan2019everybody,si2018multistage,pumarola2018unsupervised,esser2018towards,Shysheya_2019_CVPR}, 
or rendered meshes~\cite{wang2018video,liu2019neural,liu2019liquid,Sarkar2020,thies2019deferred}, 
to real images. 
Although these methods often do generate plausible images, they tend to have challenges in generalizing to different 
poses, due to complex articulations of the human body and dynamics of the clothing deformations. 
For example, methods~\cite{chan2019everybody,liu2019neural} do manage to reproduce overall body motion, 
but tend to produce severe artifacts on the hands and do not transfer facial expressions correctly.
Moreover, these methods are not capable of producing renderings from arbitrary viewpoints, which
is critical e.g. for telepresence applications.

Concurrent work to ours, SCANImate~\cite{Saito:CVPR:2021}, introduces a method for building animatable full-body 
avatars from unregistered scans, which relies on implicit representations to learn both skinning transformations 
and pose-dependent geometry correctives; interestingly, that work also discovered that localized pose 
conditioning is critical to tackle spurious correlations. 
\cite{peng2021neural} introduces a method for novel-view synthesis for human-centered
videos, which combines an articulated model (SMPL) with implicit appearance representation based on 
neural radiance fields. Although this method is capable of producing renders of arbitrary viewpoints, 
it does not allow for generalization across different poses, and is primarily tailored to short
videos. 
Neural Parametric Models~\cite{palafox2021npms} introduce a multi-identity 
parametric model for human body geometry that uses learnable implicit functions for 
shape and deformation modeling. Although promising, this method does not model 
appearance, and requires expensive optimization during inference, thus making
it unsuitable for driving with incomplete signals.

\subsection{Disentangled Representations}

The ability of a learning algorithm to discover disentangled factors
of variation in the data is considered to be a crucial property for 
building robust and generalizable representations~\cite{bengio2013representation}.
A large body of work has been focusing on building generic methods for building
disentangled representations both with~\cite{schwartz2020eyes,fadernet_NIPS2017} and without ~\cite{higgins2016beta,keyang2020unsupervised,jiang2020TVCG} supervision.

$\beta$-VAE~\cite{higgins2016beta} identified that a slight modification to the original 
VAE objective - putting a stronger weight on the prior term - can lead to automatic 
discovery of disentangled representations.
In~\cite{burgess2018understanding} authors study the properties of 
$\beta$-VAE from the perspective of the information bottleneck method, suggesting the reason behind the emergence of disentangled representations.
MINE~\cite{belghazi2018mine} provides an efficient way to compute a lower 
bound on the mutual information between two sets of variables, which can
be used as a proxy objective to encourage disentanglement.

In the context of human modeling, disentanglement has recently received
a lot of attention as a way to improve model generalization.
FaderNet~\cite{fadernet_NIPS2017} incorporates adversarial training with facial attributes for images synthesis of human faces, allowing disentanglement based on attributes.
In~\cite{schwartz2020eyes} authors propose a generative model for facial avatars 
which uses several disentanglement techniques including FaderNet to encourage a model to better make use of gaze conditioning information.
\cite{keyang2020unsupervised} introduces a generative model for 
human body meshes that uses a set of consistency losses as a way 
to separate the space of pose- and shape-based deformations.
\cite{jiang2020TVCG} achieve disentanglement for cross-identity shapes and poses by incorporating a deep hierarchical neural network.

In our settings, there are by design explicitly two groups of factors, corresponding
to the observed and missing data, respectively. 
Thus our main objective is not to discover the unknown underlying factors of variation
in an unsupervised way, but rather to encourage separation between the known observed 
factors (which are pre-defined) and the missing factors (learned space).
In Section~\ref{sec:method:missing-information} we 
discuss our approach to handling missing information in more 
detail.

\begin{figure*}[ht!]
  \centering
  \includegraphics[width=0.9\linewidth]{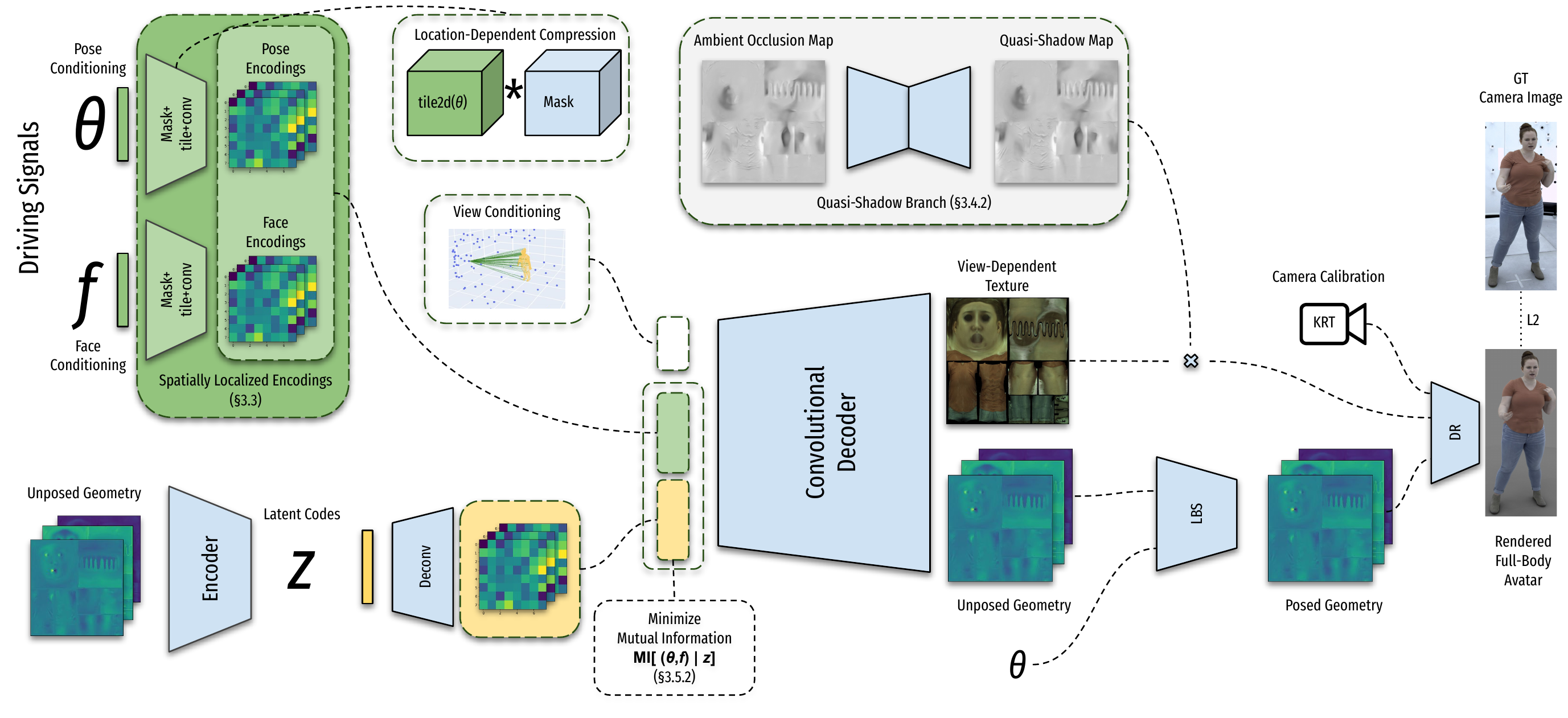}
    \caption{\textbf{General overview of the architecture}.
    The core of our full-body model is a conditional variational auto-encoder, 
    which takes as input driving signals and view direction, and outputs
    geometry and view-dependent texture. These are rendered to produce a full-body avatar. Spatially localized encoding of the driving signals helps reduce spurious correlations. Additionally, an LBS module and a quasi-shadow branch capture coarse, long-range effects. Information not present in the driving signals (e.g., clothing state) is captured by a disentangled latent code $\mathbf{z}$.}
    \label{fig:method:overview}
\end{figure*}
\section{Method}
\label{sec:method}

Our goal is to build a data-driven model for full-body avatars
that stays faithful to the driving signal, while also providing
the flexibility to generate the information that might be missing from the inputs.
For example, the driving signal for a human body might consist of sparse keypoint detections around the skeletal joints, which are insufficient to disambiguate different states of clothing, hair and facial expressions. 
Our key insight is that generalizability and controllability can
be achieved through disentanglement, while explicitly taking 
into account the information that is missing from the driving
signals.

When driving signals are sufficiently reliable, it is desirable to break 
their inter-dependencies as much as possible to achieve better 
generalization.
For human bodies, this naturally leads to the notion of 
\textit{spatially localized} control, such that, for example, a change in a 
facial expression does not have any influence on the state
of the legs. 
At the same time, driving signals often do not contain sufficient
information to fully describe free viewpoint images of the body, leading to a many-to-one mapping problem. This results in one of two cases; 1. overly smooth estimates, as the model averages over all possible unobserved states given the driving signal, or 2. overfitting to the dataset, which manifests as chance correlations between the driving signal and all unobserved factors. 
One approach to mitigate this problem is to add an additional latent code that spans the missing information space. However, as we will see in \S\ref{sec:experiments}, without special care, the model can learn to partially ignore the control signal and use the latent space to capture the full information state.
The resulting model, then, will fail to faithfully reconstruct the driving signal at test time, where the \emph{true} latent code is unobserved. 

In this work, we argue that, while indeed a latent space is useful to 
explicitly account for the missing information, it is also necessary to ensure 
that the latent space \emph{only} contains information that is \textit{not present} 
in the driving signal.
In other words, the latent space and the driving signals should be disentangled. 
As we will describe in the sections that follow, the elements of our proposed architecture are explicitly designed to achieve disentanglement through multiple and complimentary means. The result is a model that exhibits good generalization, is faithful to the driving signal during animation, and achieves good reconstruction accuracy efficiently for a representation that covers the full human body.

\subsection{Overview}

An overview of our architecture is shown in Figure~\ref{fig:method:overview}.
It takes as input the driving signals 
and viewing direction, and produces registered geometry and a view-dependent texture as output using a deconvolutional architecture. 
Together, these outputs can be used to synthesize an image through rasterization. 
There are three main components of our construction that encourage generalization through 
disentanglement, namely: spatially localized conditioning, capturing coarse long range effects, 
and information disentanglement.

\subsubsection{Spatially Localized Conditioning}
To reduce overfitting to driving signals seen during training, we employ a location-specific low-dimensional embedding. These embeddings are used to \emph{late}-condition the deconvolutional architecture, such that their \emph{footprint} in the output texture and geometry have localized spatial extent. Together, the embedding and late conditioning extract the most relevant information from the driving signal at each spatial location. This reduces the tendency to learn spurious long-range correlations which might be present in the training data. 
\subsubsection{Coarse Long Range Effects} Although localizing the effects of driving signals can reduce overfitting, there exists some long range effects that are difficult to capture with such a representation. To capture these effects without reintroducing overfitting, we identify two major sources of long range effects for human bodies and model them explicitly: rigid limb transformations and the effects of shadowing. For rigid limb transformations we employ linear-blend skinning (LBS)~\cite{magnenat1988joint,kavan2008geometric} that explicitly models rigid motion of the limbs through a composition of transformations along an articulated tree structure that spans the entire body. We thus decouple our decoder into using joint angles via LBS, and producing correctives in the form of a displacement map generated using the deconvolutional architecture. To handle long range shadowing effects, we explicitly compute an ambient occlusion map using the LBS-generated geometry, and pass it through a UNet~\cite{unet_miccai2015} to produce a gain map that is applied on the output texture. The ambient occlusion map serves as a substitute for a shadow-map since our capture space has roughly uniform illumination. Together, the LBS and shadow branches complement the localized conditioned described above, so that major long-term effects are accounted for while maintaining good generalization performance. 
\subsubsection{Information Disentanglement} The main motivation for introducing a latent space is to capture
information not contained in the driving signal, yet necessary to fully explain image evidence.
Unlike the driving signal, where avoiding overfitting is primary, variations that have no supporting evidence during animation require the strongest possible prior to ensure compelling imputation. Thus, we use a vector-space representation at the bottle-neck of our architecture for the latent space that has spatial support over the entire output. To ensure drivability, we encourage the latent codes to contain the least amount of information about the driving signal as possible by employing disentanglement strategies during training.
We train all these components of our model jointly using a Conditional Variational Auto Encoder (cVAE)~\cite{sohn2015learning} that has been used effectively to produce well-structured latent spaces in human-centric datasets~\cite{fadernet_NIPS2017,lombardi2018deep}. For supervision, we directly use multi-view images, which we achieve by applying differentiable rendering~\cite{liu2019softras} to generate synthetic images for a direct comparison with the ground truth.

\subsection{Variational Autoencoder}
\label{sec:VariationalAutoencoder}

The core of our model is a conditional variational auto-encoder (cVAE),
consisting of an encoder $E(\cdot | \bW_E)$, and a decoder 
$D(\cdot | \bW_D)$, both parameterized as convolutional neural networks, with weights $\bW_E$ and $\bW_D$, respectively. The cVAE is trained end-to-end to reconstruct images of a subject captured from a multi-view camera rig. 
Our system presumes the availability of a registered mesh, $\bG_i \in \mR^{N_v \times 3}$, for each frame $i$ that we acquire using LBS-based tracking with surface registration~\cite{gall2009motion}, followed by Laplacian deformation to 3D scans for better registration~\cite{botsch_tvcg2008,Sorkine04sgp}. The LBS joint angles, $\btheta_i \in \mR^{100}$, as well as 3D facial keypoints, $\bbf_i \in \mR^{N_f}$, are the driving signals that we use to condition our model. For settings where 3D face keypoints are not available, such as in the experiment with a VR headset in \S\ref{sec:experiment_driving}, we use the latent codes of the personalized facial model from~\cite{wei2020VRFace} instead. 
Finally, we note that the goal of our cVAE training is to produce a decoder that we can use to animate an avatar from driving signals at test time. To this end, the role of the encoder is strictly to facilitate learning and ensure a well structured latent space. It can be discarded once training is complete.

During training, the encoder takes as input geometry, $\bG_i$, that has been unposed using LBS, and 
produces parameters of a Gaussian distribution, $\mathcal{N}$,
from which a latent code $\bz \in \mR^{512}$ is sampled:
\begin{equation*}
\label{eqn:variational}
    \bmu_i, \bsigma_i \leftarrow E(\bG_{i} | \bW_E) \; , \;  \bz_i \sim \mathcal{N}(\bmu_i, \bsigma_i^2).
\end{equation*}
Here, $\bG_i$ is first rendered to a position map in UV space before being passed to the convolutional encoder.
The reparameterization trick~\cite{kingma2013auto} is used to ensure differentiability of the sampling process. 
Given the latent codes $\bz_i$, driving signals ($\btheta_i, \bbf_i$) and a viewpoint $\bv_i \in \mR^3$, the decoder produces reconstructed geometry $\tilde{\bG}_i$ and view-dependent texture $\tilde{\bT}_i$:
\begin{equation*}
    \tilde{\bG}_i, \tilde{\bT}_i \leftarrow D(\btheta_i, \bbf_i, \bz_i, \bv_i | \bW_D).
\end{equation*}
$\tilde{\bG}_i$ is then passed to the LBS module to produce the final geometry, and $\tilde{\bT}_i$ is multiplied 
by a quasi-shadow map to produce the final texture. 
The generation of the quasi shadow map is detailed in \S\ref{sec:SelfShadowing}. Differentiable rasterization~\cite{liu2019softras} is used to render an image using the output geometry and texture, which is then compared with the ground truth image using an L2-error. 
Along with other supervision and regularization losses described in \S\ref{sec:TrainingDetails}, the system is trained 
end-to-end, solving for the cVAE parameters $\bW_E$ and $\bW_D$.

Once the model is trained, animation is performed by taking the driving signals ($\btheta, \bbf$) 
e.g. estimated from an external sensor, imputing the latent code $\bz$ (e.g. by sampling or 
employing a temporal model), and then synthesizing 
the geometry and texture by running the decoder $D(\btheta_i, \bbf_i, \bz_i, \bv_i | \bW_D)$ followed by rasterization. 
The reason for separating conditioning signal between ($\btheta, \bbf$) and $\bz$ comes 
from the fact that, in practice, access to the \textit{complete} conditioning signal 
is available only during training.
%
This stems from differences in their capture setup: training data 
collection often allows for 
sophisticated multi-camera rigs whose data can be processed in post,
whereas during animation, the capture system is typically more constrained and requires real-time processing.
Our specific choice of ($\btheta, \bbf$) in this work is informed by attributes that can be reliably inferred from limited sensing setups in real-time, such as keypoints and skeleton joint angles from a simple stereo camera pair~\cite{tan2020efficientdet,gall2009motion}.
Ultimately, this means that the information contained in the driving signal is often insufficient
to fully describe the output, and we need to introduce $\bz$ that describes 
the remaining part of the signal, 
so as to avoid model overfitting or smoothing out the results.
In Section~\ref{sec:experiments}, we provide a comparison with a baseline version
of the model that does not use a latent space.

\subsection{Spatially Localized Driving Signals}
\label{sec:SpatiallyLocalizedDrivingSignals}

\begin{figure}[ht!]
  \centering
     \includegraphics[width=1.0\linewidth]{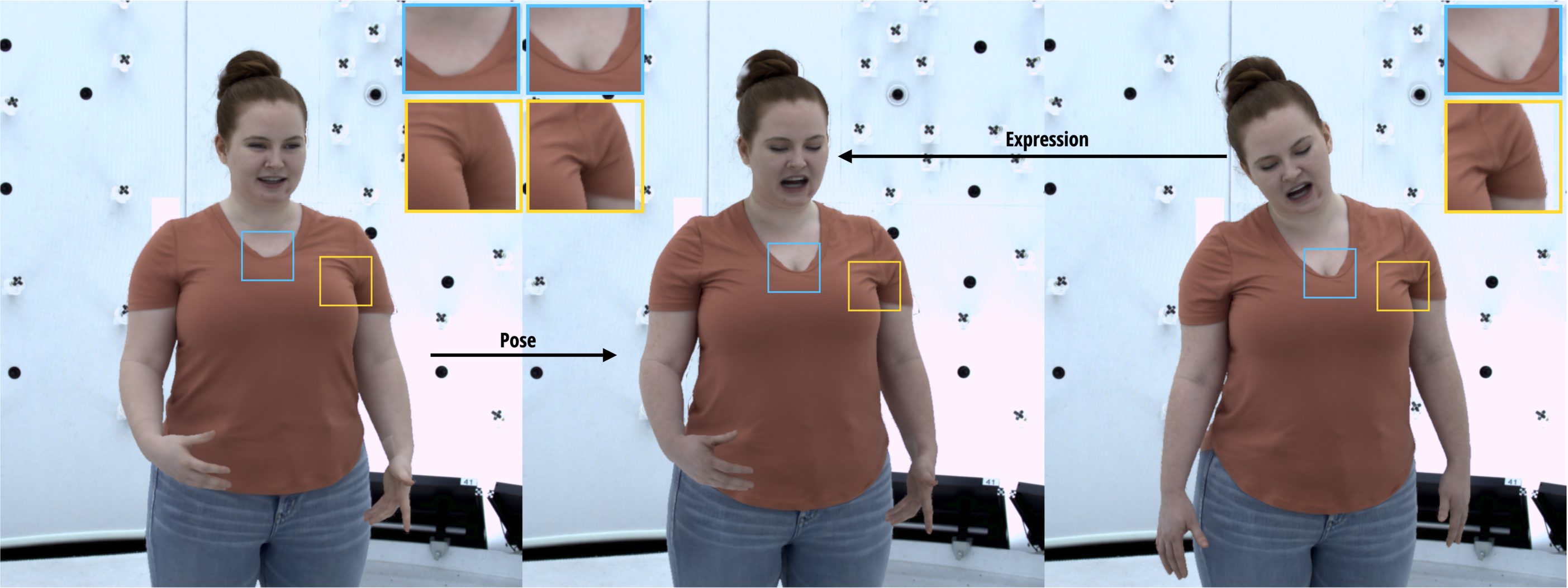}
    \caption{\textbf{Spurious Correlations}. 
    A naive model learns to associate clothing and shadows 
    facial expression. Combining the pose from the sample on the left, and the facial
    expression from the image on the right produces an unrealistic
    sample where pose-dependent effects are fully controlled by the facial 
    expressions (e.g. the collar and armpits, which should be independent of expression).
    }
    \label{fig:method:spurious-correlations}
\end{figure}

For a model to be truly drivable, it has to generalize well. That is, it should
produce realistic outputs for all real combinations of the driving signals.
We are focusing on building data-driven models, and thus a naive approach to achieving
generalization would be to collect an exhaustive amount of data that would 
sufficiently cover the space of variations.
Unfortunately, even for personalized full-body models with a single attire, collecting a dataset
that can cover all possible combinations of body posture, facial expressions and hand gestures is intractable due to the combinatorial explosion of part variations. 
%
A common approach is to instead capture range-of-motion data, with the aim of spanning the full range of each body part with the hope that the model can learn to factorize these parts accordingly. 
%
However, if one considers highly-expressive models, such as deep ConvNets, 
relying on such limited data can lead to situations where the model 
discovers spurious correlations and thus learns to encode capture-specific
dependencies that are not present in other sequences.
For example, if during the capture of hand gestures the subject keeps the same facial expression throughout, there is no incentive for the model \textit{not to} learn an association between that facial expression
and the hand gestures. An example of this phenomenon occurring on real data is shown in Figure~\ref{fig:method:spurious-correlations}.

However, if we assume that the driving signal is sufficiently reliable,
and it is not necessary to capture correlations between all of its
individual components, we can build a model in a way that
encourages decoupling between those individual components.
%
For the human body this would correspond to the fact that spatially distant 
parts (e.g. fingers on different hands) can move completely independently from each 
other.
Similar intuition has been applied for human geometry in STAR~\cite{STAR:2020}, which proposes a localized version 
of pose correctives aimed at reducing spurious correlations: instead of \textit{global} blendshapes as in SMPL~\cite{bogo2016keep}, 
the authors propose \textit{local} blendshapes with a non-linear blending scheme.
Note that this reasoning is valid primarily for geometrical deformations 
and appearance effects which are \textit{local}, but does not hold for global 
ones, such as shadowing, and one has to model them separately, as
we will discuss in \S\ref{sec:SelfShadowing}.

\begin{figure}[ht!]
\centering
\begin{tabular}{@{}c@{}c@{}c@{}}
    \includegraphics[trim={0cm 3cm 0cm 7cm},clip,width=0.16\textwidth]{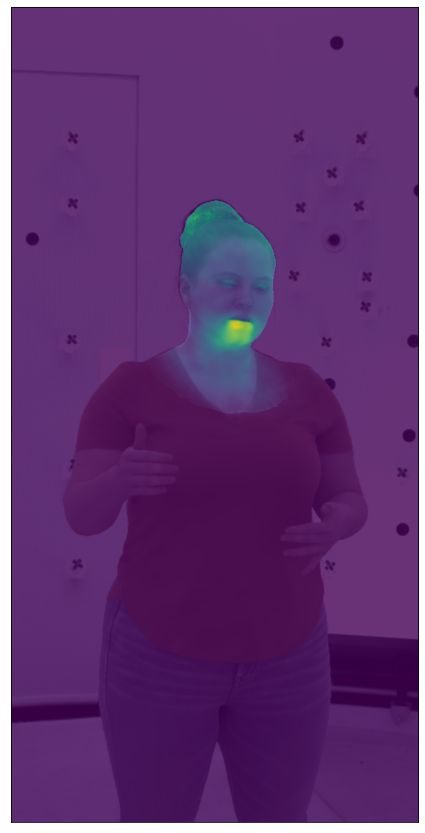}&
    \includegraphics[trim={0cm 3cm 0cm 7cm},clip,width=0.16\textwidth]{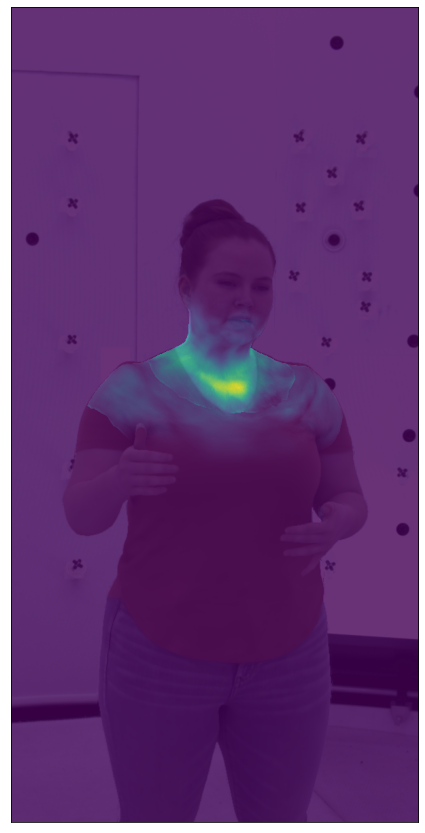}&
    \includegraphics[trim={0cm 3cm 0cm 7cm},clip,width=0.16\textwidth]{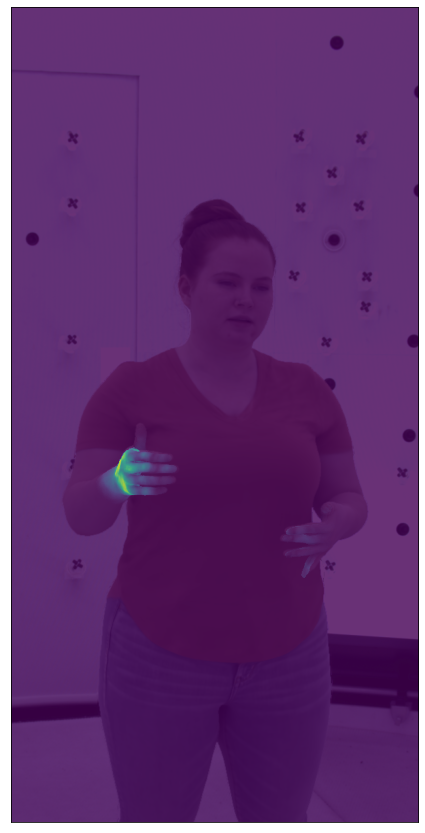}\\
    \end{tabular}        
    \caption{\textbf{Effects of localized representations}. 
    Heatmaps indicate the areas with largest changes in the output
    of our model when varying a single control variable.
    From left to right: face, neck, right wrist.}
    \label{fig:method:masks}
\end{figure}

In this work, we rely on the structure of our decoder network to achieve conditional spatial independence given our driving signal. 
Specifically, our decoder is a fully-convolutional network which takes as inputs
several encoding maps, $\be_* \in \mR^{N_{*}\times32\times32}$,
one for each driving source, $* \in \{\btheta, \bbf\}$, where:
\begin{equation*}
\be_* = \mathrm{proj}( \bM_* \odot \mathrm{tile2d}(*, 32, 32) \ | \ \bW_*).
\end{equation*}
Here, $\mathrm{tile2d}(\bx, h, w)$ is an operation that repeats a given
vector $\bx$ into a $h \times w$ feature map,
and $\mathrm{proj}(\cdot | \bW_{\bx})$ is a projection operation
that applies a location-dependent compression at each
point of the input feature map.
%
It is implemented using a two-layer 1x1-convolutional network with untied biases.
%
We also apply a binary mask $\bM_{x} \in \{0, 1\}^{100\times32\times32}$, 
where each channel roughly defines the region of influence for each
parameter, which we define using downsampled LBS skinning weights and capture the local effects of each joint angle.
%
In principle, these masks can also be learned, 
but we did not observe improved results by doing so in our experiments.
Figure~\ref{fig:method:masks} visualizes the effects of
using localized driving signals, where varying one of them
leads to spatially (and semantically) coherent changes in the corresponding 
region of the output.


\subsection{Coarse Long Range Effects}
\label{sec:method:CoarseLongRangeEffects}

Some common driving signals used for human body animation have long range effects that couple distant locations on the body. Explicitly localizing the effects of the driving signal, as described previously in \S\ref{sec:SpatiallyLocalizedDrivingSignals}, can therefore limit the model's ability to reconstruct real data with high accuracy. But naively reintroducing global influence can lead to over-fitting. As such, we identify two major sources of long-range effects in human body appearance, rigid limb motion and self-shadowing, and propagate these effects spatially using parametric forms that have been shown to generalize well to novel poses.

\subsubsection{Rigid Limb Motion}
\label{sec:RigidLimbMotion}

Similar to some existing work~\cite{loper2015smpl,SCAPE05,STAR:2020}, 
our geometric model composes the localized model, described in \S\ref{sec:SpatiallyLocalizedDrivingSignals}, with an explicit skeleton model based on LBS, which captures large rigid transformations of limbs:
\begin{equation*}
    \bar{\bG} = LBS(\btheta, D_G(\btheta, \bbf, \bz) + \bar{\bG}_T).
\end{equation*}
Here, $D_{G}$ is the geometry branch of the decoder, $\bar{\bG}_T$ is the \textit{template} mesh in canonical pose, and $\bar{\bG}$ is the final geometry after posing\footnote{We have dropped the frame index $i$ from $\btheta_i$, $\bbf_i$ and $\bz_i$ to reduce clutter.}. 
In particular, given the template and the pose parameters, each vertex $\bx^i \in \mR^3$ of $\bar{\bG}$ is 
computed as: 
\begin{equation*}
  \bx^i =\sum_{j} w_{ij} \cdot (\bR_j \cdot \bx^i_T + \bt_j) \; , 
\end{equation*}
where $w_{ij}$ are the pre-defined skinning weights, which describe the influence of 
$j$-th joint on the $i$-th vertex $\bx^i$, and $\bR_j$, $\bt_j$ are per-joint 
transformation parameters, which are computed from $\btheta$ via forward
kinematics.

The LBS model has been demonstrated to capture large coarse motion of the human body, linking distant locations via a kinematic chain. Although many of its limitations are well known, e.g., the candy wrapper effect~\cite{kavan2008geometric} and inability to model more complex nonlinear effects~\cite{jacobson2011stretchable,loper2015smpl}, these deficiencies tend be localized and have been shown to be addressable by using localized corrective models~\cite{STAR:2020}. Our approach builds on these prior findings, complementing LBS with the localized model described previously in~\S\ref{sec:SpatiallyLocalizedDrivingSignals}.

\subsubsection{Self-Shadowing}
\label{sec:SelfShadowing}

Whereas rigid limb motion accounts for much of the long range geometrical effects, self-shadowing accounts for the majority of long-range appearance effects. This is particularly the case during self-interaction, where limbs touch, or are in close proximity to each other. Explicitly accounting for these effects frees up the localized model in \S\ref{sec:SpatiallyLocalizedDrivingSignals} to focus on the remaining sources of appearance variations, which tend to be more localized, such as skin wrinkles, clothing shifts, and view-dependent effects. 

Physically correct estimation of self-shadowing is an extensive process~\cite{hillsiggraph2020_shading}. It requires an accurate model of the lighting environment and geometry that may not be readily available. In our case, this would require first running the whole geometry branch before an estimate of shadowing can be used to inform the texture branch, which limits the use of parallelism for achieving fast decoding. Instead, in this work we compute a fast approximate model for depicting environment-occlusion, and then learn the shadowing effect implicitly.

\begin{figure}[ht!]
  \centering
    \begin{tabular}{@{}c@{}c@{}}
    \includegraphics[trim={0cm 6cm 0cm 3cm},clip,width=0.24\textwidth]{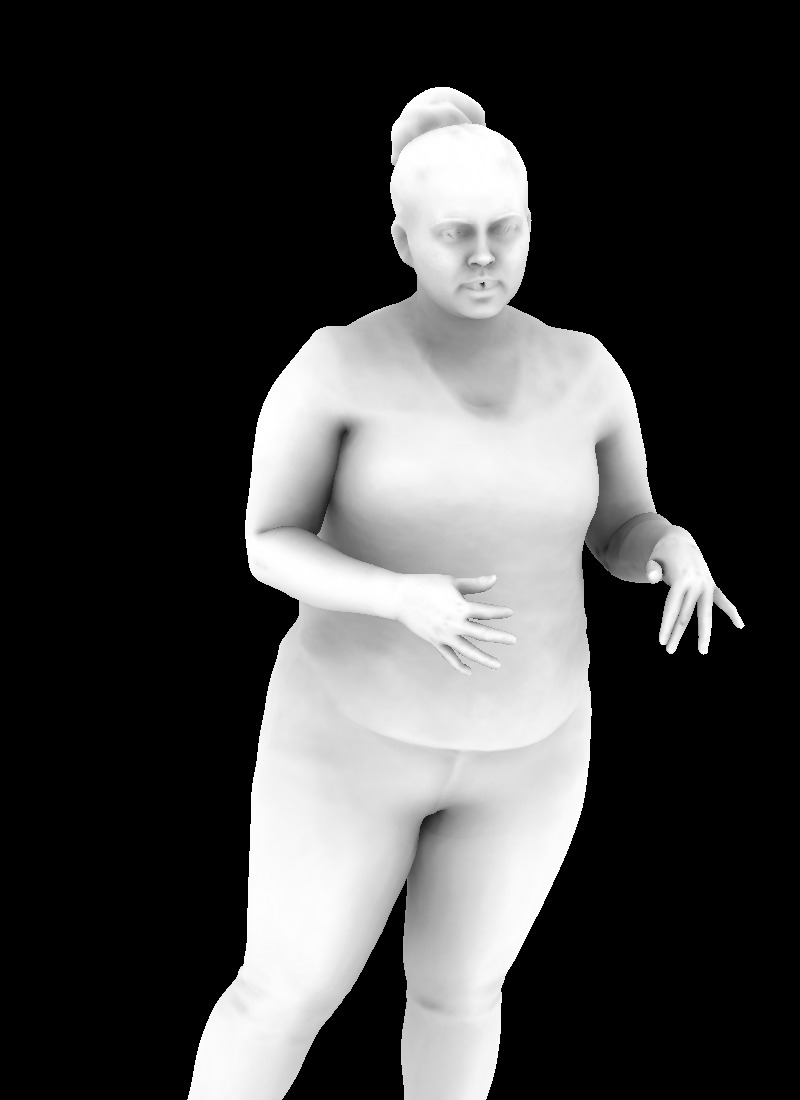} &
    \includegraphics[trim={0cm 6cm 0cm 3cm},clip,width=0.24\textwidth]{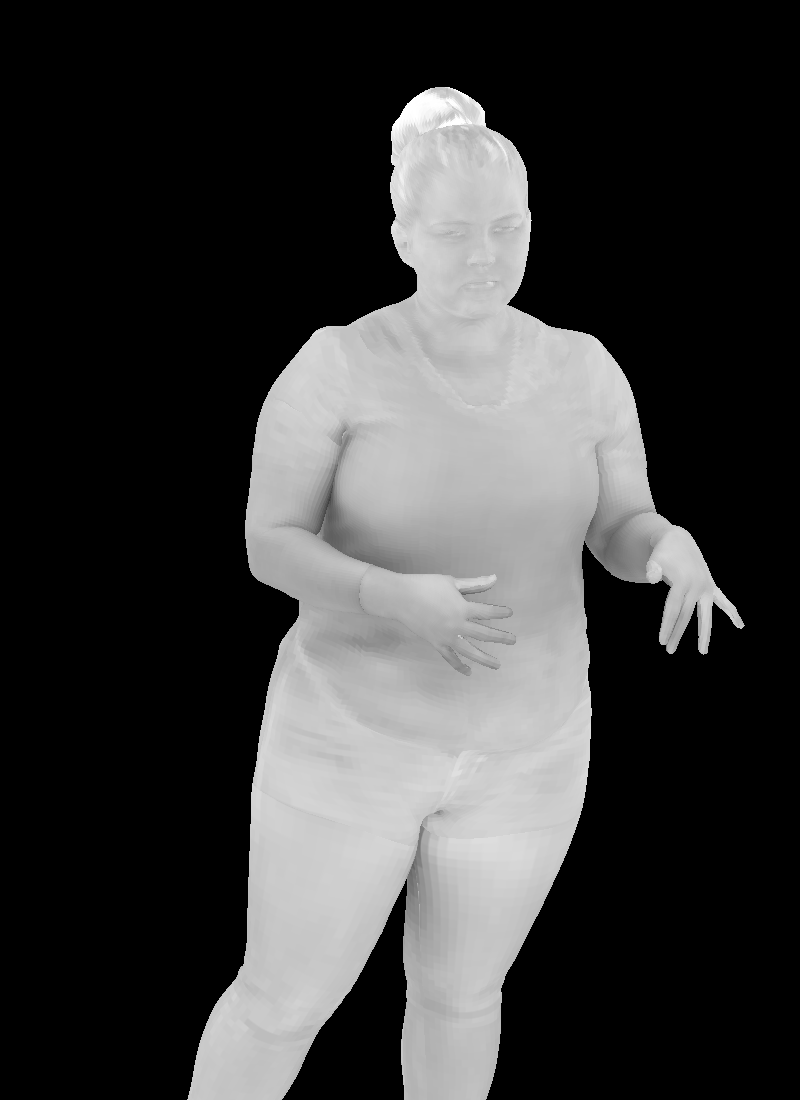}
    \end{tabular}  
    \caption{\textbf{Shadow branch inputs and outputs}. 
    The left image is an example of a conditioning signal for the network, 
    and the right image is the prediction of our network on a test
    sample. Note that even though there is no explicit
    supervision provided to the shadow branch, it naturally
    learns to capture approximate global shadowing.
    }
    \label{fig:method:ao-examples}
\end{figure}

Ambient occlusion (AO) captures environment visibility at each location of the body. It is a strong feature for self-shadowing in a static environment that can be computed efficiently~\cite{miller_AO_SIGGRAPH1994}. Additionally, for roughly uniform illumination conditions, as in our case, dependence on the global pose with respect to the environment is unnecessary. For computing an AO map, we use the LBS-posed template geometry as it is efficient to compute and does not require the full evaluation of the geometry branch to generate. Although this reduces the precision of the AO map, our goal here is to capture coarse long-range appearance effects, and the differences resulting from using the posed template geometry tend to be localized. This AO map is then passed through the shadow branch; a neural network with a UNet architecture~\cite{unet_miccai2015} that produces a quasi-shadow map. The result is multiplied with the output of the texture branch to produce the final texture. Note that the shadow branch is not supervised directly. Rather, it is used as an inductive bias in cVAE, where its supervision is implicit through a comparison of the final rendered image with the ground truth capture (see \S\ref{sec:VariationalAutoencoder}). 
In practice, we found that it is sufficient to produce a low-resolution
quasi-shadow map (4x less compared to the texture) to obtain plausible
soft shadows. 
%
An example of input and output of the shadow branch is given in Figure~\ref{fig:method:ao-examples}.

Owing to its physically inspired formulation, AO computation generalizes to novel geometry generated through LBS, which has already demonstrated its ability to generalize to new poses, given that reliable joint angles can be acquired from the driving signals. The UNet architecture has been shown to exhibit good generalization~\cite{unet_miccai2015}, especially when its input and outputs differ mostly by local changes, as in our case, by allowing it to rely mostly on shallow skip connections\footnote{For non-uniform lighting environments, the generalization performance of this network may be affected since the input AO map would differ from the output shadow map more severely. One solution to this might be to perform a full shadow map calculation that the UNet would then only need to refine. We leave that investigation as future work.}. As with LBS for geometry, the shadow branch frees up the texture branch to focus on local appearance effects since the majority of long range effects are already accounted for through the shadow maps.

\subsection{Handling Information Deficiency}
\label{sec:method:missing-information}

In this section, we describe our approach for handling the problem of information deficiency
during animation that was outlined in \S\ref{sec:VariationalAutoencoder}.
Our main assumption is that one can identify an \textit{explicitly} defined 
set of control variables, such as $\btheta$ and $ \bbf$, that can be reliably estimated 
during both training and animation. 
The choice for driving signals is mainly driven by the application and the sensors it affords. 
A common set of signals that can be acquired for driving include the full-body pose and facial keypoints estimated from an cameras placed in the scene, or, when deploying to a VR-based application, on a headset. 


Only providing the model with body joints and face keypoints is, however, insufficient,
as they do not contain all the information required to explain the full appearance of a body in motion, such as the specific state of clothing, hair and internals of the mouth for a given image. Due to the high dimensionality of the inputs as well as the high capacity of neural networks, directly regressing the output from these signals alone typically overfits to the data, where it learns correlations between the driving signal and outputs that are not generalizable, e.g. by 
remembering a certain shadow or wrinkle combination.

\begin{figure}[ht!]
  \centering
   \includegraphics[width=1.0\linewidth]{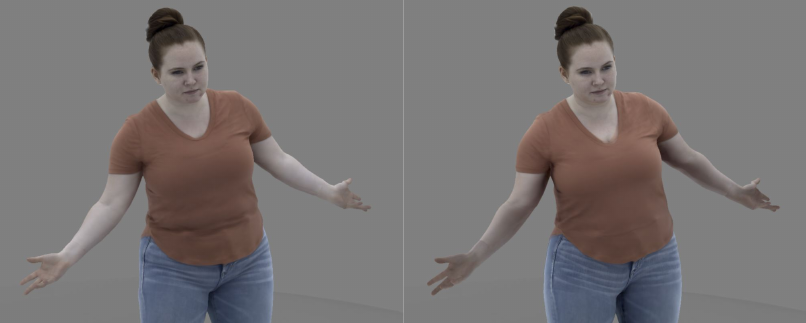}
    \caption{\textbf{The need for disentangling}. 
    A model with a latent space but no disentangling mechanism produces reasonable 
    reconstruction (left) but ignores the pose conditioning, which leads
    to severe artifacts when driving (right). For example, incorrect shading on the arms, ghosting artifacts on the collar and overly smoothed clothing wrinkles.
    }
    \label{fig:method:no-disentangling-artifacts}
\end{figure}

To address this problem, a naive solution would be to simply introduce an additional latent 
variable $\bz$ that contains the remaining information about the outputs,
e.g. by encoding that latent code from the outputs.
However, this approach leads to another problem; there is no
guarantee that $\bz$ is \textit{independent} of the driving signal. 
In practice, it tends to retain some 
information about ($\btheta, \bbf$), 
which leads the decoder to 
ignore either of the two variables, or worse, learn
to "spread" the information between the two,  
eliminating the ability to remain faithful to the driving signal during animation\footnote{A related problem is that of posterior collapse~\cite{aliakbarian2019mitigating}, where a high capacity decoder or strong conditioning signal can lead to the latent space being ignored. This scenario is more common in temporal auto-regressive models, where the history used for conditioning can be highly predictive of the current state. We did not observe this behavior in our setting, possibly due to the conditioning signal not being strong enough or exhibiting a complex relationship with the output, where the information path from encoder to decoder is learned more easily. }.
One example of such a failure mode is illustrated in Figure~\ref{fig:method:no-disentangling-artifacts}.

In this work, we propose three strategies to \textit{disentangle} the two sets of variables, which one can select between to best match the characteristics of the data at hand. 

\subsubsection{Variational Preferencing}
\label{sec:method:variational_preferencing}

cVAEs have a built-in mechanism for disentangling conditioning variables from the latent space. This stems from the propensity of a VAE with a factorized Gaussian posterior to push a portion of its latent dimensions to become uninformative in order to better minimize its KL-regularized loss. In effect, it squeezes information at the input to the encoder into a lower-dimensional subspace within its latent space, the dimensionality of which can only be reduced as the KL weight increases. 

In cVAE, when the conditioning variable shares some information with the input to the encoder, there is a preference to let the decoder use the \emph{copy} of that information contained in the driving signal, so as to allow it to be squeezed out of the latent space, thereby achieving lower KL-divergence while still providing the decoder with information necessary for good reconstruction. This results in disentanglement between the latent space and conditioning variables. 

By employing a cVAE architecture, our model already exhibits a built-in disentangling mechanism between the latent space and the driving signal which is used to condition the decoder. However, VAE's disentangling mechanism is not perfect. Architecture choices for the encoder and decoder, as well as optimization strategies and even the characteristics of the data can either encourage or impede disentanglement. Furthermore, increasing the KL-weight as a way of achieving better disentanglement can negatively impact reconstruction accuracy as the effective dimensionality of the latent space becomes further reduced. 

\subsubsection{Mutual Information Minimization}
\label{sec:method:mutual-information-minimization}

A more direct approach to \textit{disentangle} the two sets of variables is by minimizing the mutual information (MI) between them:
\begin{equation*}
    \mathcal{I}(\bz, \bc) = \mathrm{KL}(p (\bc, \bz) || p (\bc) p (\bz) ) \;,
    \label{eq:method:mi}
\end{equation*}
i.e. a KL-divergence between the joint distribution and the factorized one,
where $\bc = (\btheta, \bbf)$.
Since directly computing MI is intractable, we use MINE~\cite{mattei2019miwae}  - a generic data-driven approach 
for estimating the mutual information between random variables.
%
%
It introduces a parametric approximation to MI through a "statistics network" implemented using a deep neural network:
\begin{equation*}
    m \leftarrow f(\bz, \bc) \; ,
\end{equation*}
where $m \in \mR$ is a scalar that is high when the mutual information between $\bz$ and 
$\bc$ is high, and low otherwise. 
%
This statistics network $f(\cdot)$ is trained 
using the following loss function with each mini-batch: 
\begin{equation}
\mL_{\mathrm{mi}} = - (\frac{1}{B}\sum_{b=1}^{B}  f(\bc_b, \bz_b) - \log \sum_{b=1} \frac{1}{B} e^{f(\bc_b, \hat{\bz}_b)}) \;, 
\label{eq:label:mi-estimate}
\end{equation}
where $B$ is the size of a mini-batch, and $\hat{\bz}_b$ is a shuffled, or randomly sampled, embedding.
Thus, the network is trained to maximize the difference in $f$-function scores between when the data is paired vs. randomized. 
This results in a (biased) estimate of MI, which can be used to evaluate how independent $\bc$ and $\bz$ are.
The loss in Eq.\ref{eq:label:mi-estimate} can be used to define an
adversarial term $\mL_{\mathrm{dis}} = - \mL_{\mathrm{mi}}$, which
we add to
the full model training objective described in \S\ref{sec:TrainingDetails}. 
This way, our cVAE learns to reconstruct the training data while disentangling the latent space from the driving signals.

\subsubsection{Perturbation Consistency}
\label{sec:method:perturbation-consistency}

When the driving signal has a direct physical interpretation with respect to the model's output, there is an even more direct way to encourage disentanglement. For example, when the driving signal comprises a set of keypoints on the surface of the body, it may be possible to define direct correspondence with vertices on the output geometry. In this case, disentanglement between the latent code, $\mathbf{z}$, and the driving signal, $\mathbf{c}$, can be achieved using a perturbation consistency loss:
\begin{equation}
    \mL_{\mathrm{pc}} = \mathbb{E}_{p(\mathbf{z})}\left [ 
     \sum_{k \in \mathcal{C}}
    \left \|
        \mathbf{c}_k - \mathbf{S}_k \ D(\mathbf{\mathbf{z}, \mathbf{c} | \mathbf{W}_D)}
    \right \|^2
    \right ],
\end{equation}
where $p$ is the prior over the VAE's latent space, $\mathcal{C}$ is the set of driving signal components that admit a correspondence with the output from the decoder, $D$, and $\mathbf{S}_k$ is a selection matrix that picks the corresponding elements in the output. In practice, the expected value is approximated by a discrete sampling of $\mathbf{z}$ for each minibatch during training. If $\mathbf{z}$ contains information about $\mathbf{c}$, a random sample will perturb the corresponding elements of the decoded output away from $\mathbf{c}$. Similarly, if $\mathbf{c}$ and $\mathbf{z}$ are independent, modifying $\mathbf{z}$ will have little impact on $\mathbf{c}$-corresponded outputs of $D$, so long as $D$ can reconstruct the data well. 

Although the perturbation consistency loss is the most direct way of encouraging independence between the latent space and driving signal, it is not always possible to define direct correspondence between all elements of the driving signal and the output of the decoder. In practice, we find a good heuristic is to start by relying simply on variational preferencing, and if that is not enough, add mutual-information minimization, and finally to add perturbation consistency whenever it is available.


\subsection{Training Details}
\label{sec:TrainingDetails}

We train all our models with Adam, using an initial learning rate of 1.0e-3, and batch size 8.
All models are trained until convergence on 8 NVidia V100 GPU with 32GB 
RAM.
For full body models, training takes approximately $1.5-2$ days.
We optimize the following composite loss:
\begin{equation}
\mL = 
\lambda_{\mathrm{I}} \mL_{\mathrm{\mathrm{I}}} + 
\lambda_{\mathrm{M}} \mL_{\mathrm{\mathrm{M}}} + 
\lambda_{\mathrm{lap}} \mL_{\mathrm{lap}} + 
\lambda_{\mathrm{G}} \mL_{\mathrm{G}} + 
\lambda_{\mathrm{KL}} \mL_{\mathrm{KL}} +
\lambda_{\mathrm{dis}} \mL_{\mathrm{dis}} \; ,
\end{equation}
where $\mL_{\bI}$ is the inverse rendering losses for images 
$\bI$ and foreground masks $\bM$
\begin{equation*}
\mL_{\bI} = ||\bI_{\mathrm{gt}} - \hat{\bI}||_1 \;,
\mL_{\bM} = ||\bM_{\mathrm{gt}} - \hat{\bM}||^2_2 \;,
\end{equation*}
$\mL_{\mathrm{G}}$ is the per-vertex tracked mesh loss:
\begin{equation*}
\mL_{\bG} = ||\bG_{\mathrm{gt}} - \hat{\bG}||^2_2 \,
\end{equation*}
$\mL_{\mathrm{lap}}$ is the Laplacian loss with respect to the ground truth tracked mesh,
which encourages smoothness
\begin{equation*}
\mL_{\mathrm{lap}} = ||L(\bG_{\mathrm{gt}}) - L(\hat{\bG})||^2_2 \;,
\end{equation*}
where $L(\cdot)$ is the mesh Laplacian operator. $\mL_{\mathrm{KL}}$ is the variational term,
which is a standard VAE KL-divergence penalty~\cite{kingma2013auto},
and $\mL_{\mathrm{dis}}$ is the disentangling loss as described in \S\ref{sec:method:missing-information}.
In practice, we only use mesh supervision $\mL_{\mathrm{G}}$ during
the first 2000 iterations so as to provide a reasonable
initialization for the model.
After that, the inverse rendering loss on images and masks further improves 
the shape and correspondence, as the initial tracked mesh may exhibit artifacts 
due the tracking challenges of human body surface. 
We find that this two phase procedure leads to improved estimates of shape 
and correspondence, which boosts the accuracy at which our decoder can 
reconstruct the images.

\section{Experiments}
\label{sec:experiments}

\begin{figure}[t]
    \centering
    \includegraphics[width=\linewidth]{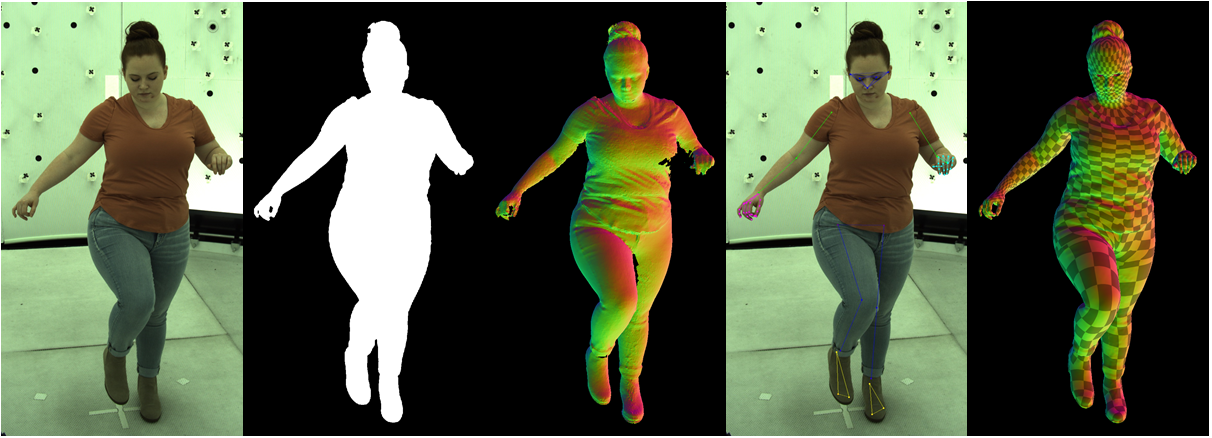}
    \caption{\textbf{Data Processing Pipeline.} From left to right: input image, foreground body mask, reconstructed 3d shape, detected keypoints, registered surface mesh.}
    \label{fig:data-processing}
\end{figure}

\begin{figure}[t]
    \centering
    \includegraphics[width=\linewidth]{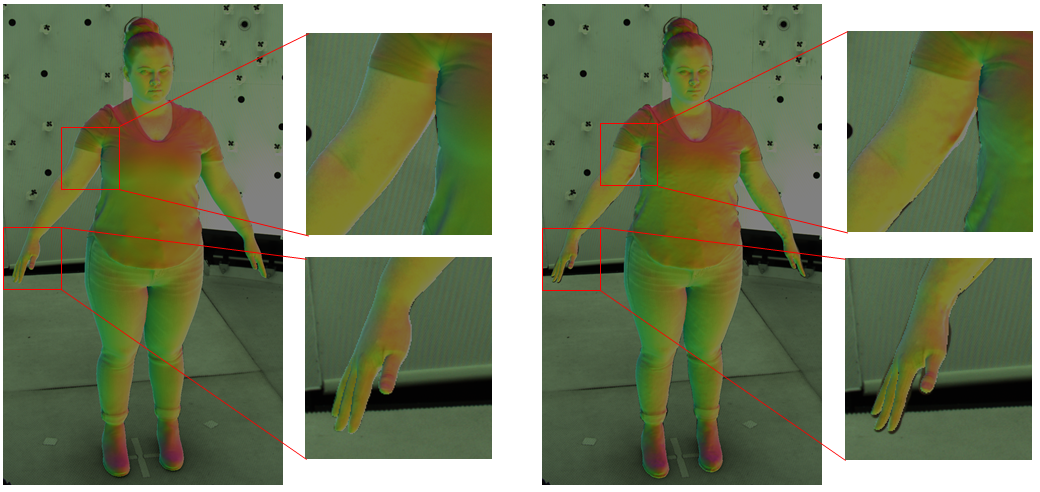}
    \caption{\textbf{Geometry Improvement by Inverse Rendering.} Left: our result after inverse rendering; right: initial tracked mesh from data processing.}
    \label{fig:inverse-rendering}
\end{figure}

\begin{figure*}
  \begin{tabular}{@{}c@{}c@{}c@{}c@{}}
\includegraphics[trim={2.5cm 2cm 2.5cm 6cm},clip,width=0.24\textwidth]{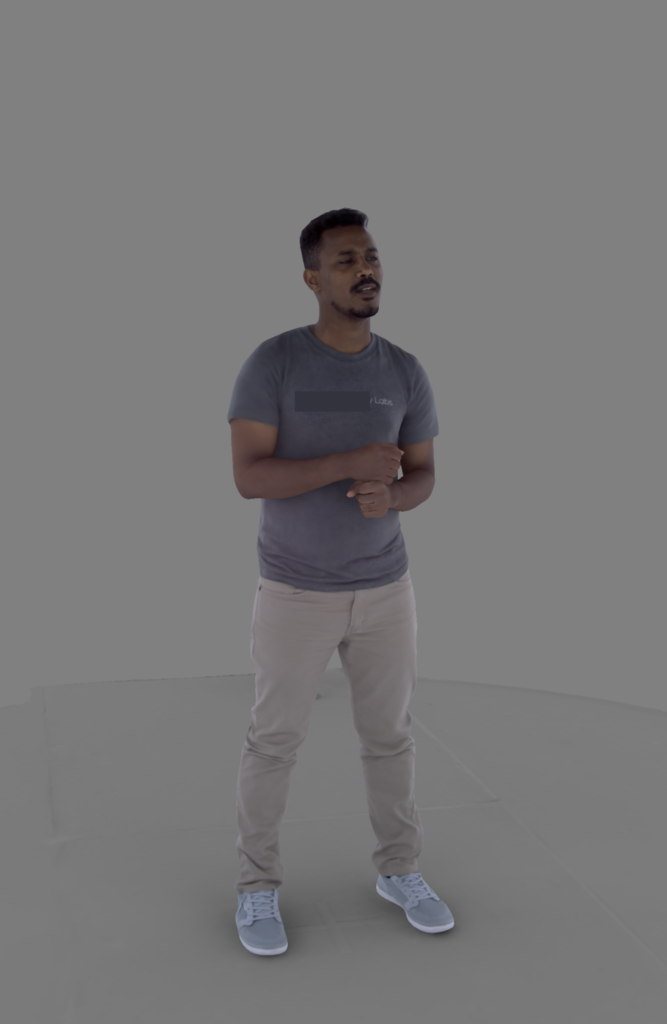}&
\includegraphics[trim={2.5cm 2cm 2.5cm 6cm},clip,width=0.24\textwidth]{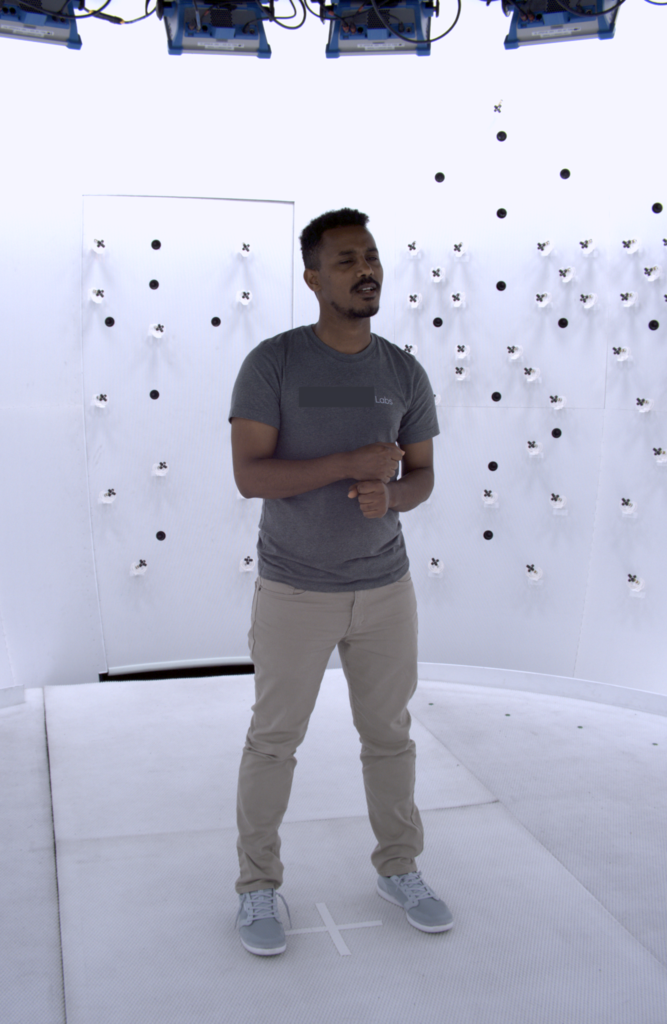}&
\includegraphics[trim={2.5cm 2cm 2.5cm 6cm},clip,width=0.24\textwidth]{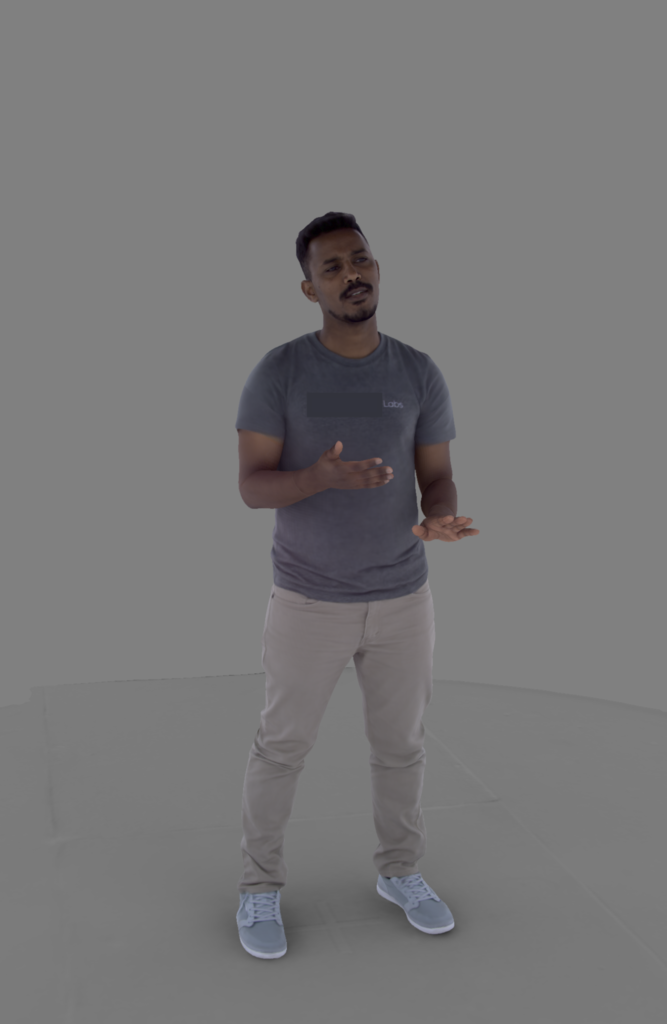}&
\includegraphics[trim={2.5cm 2cm 2.5cm 6cm},clip,width=0.24\textwidth]{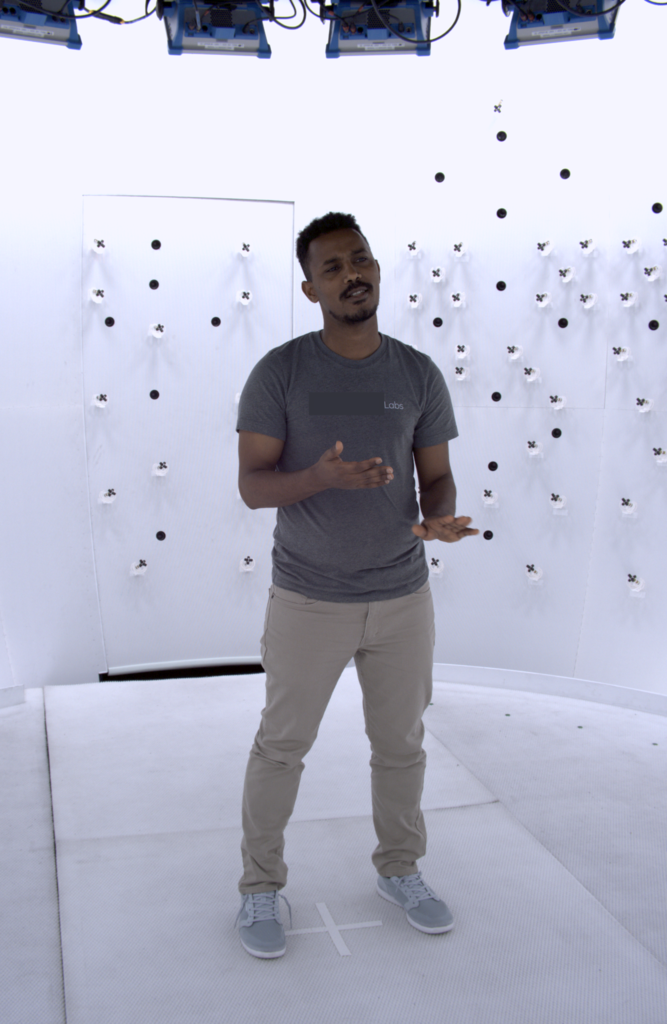}\\
\includegraphics[trim={4cm 2cm 2.5cm 8.5cm},clip,width=0.24\textwidth]{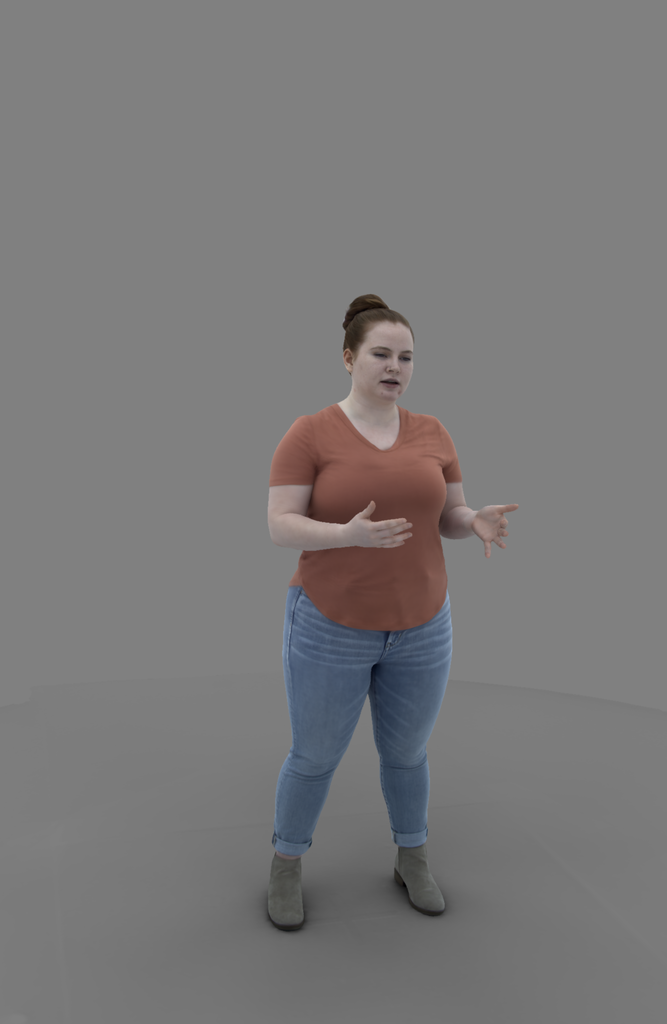}&
\includegraphics[trim={4cm 2cm 2.5cm 8.5cm},clip,width=0.24\textwidth]{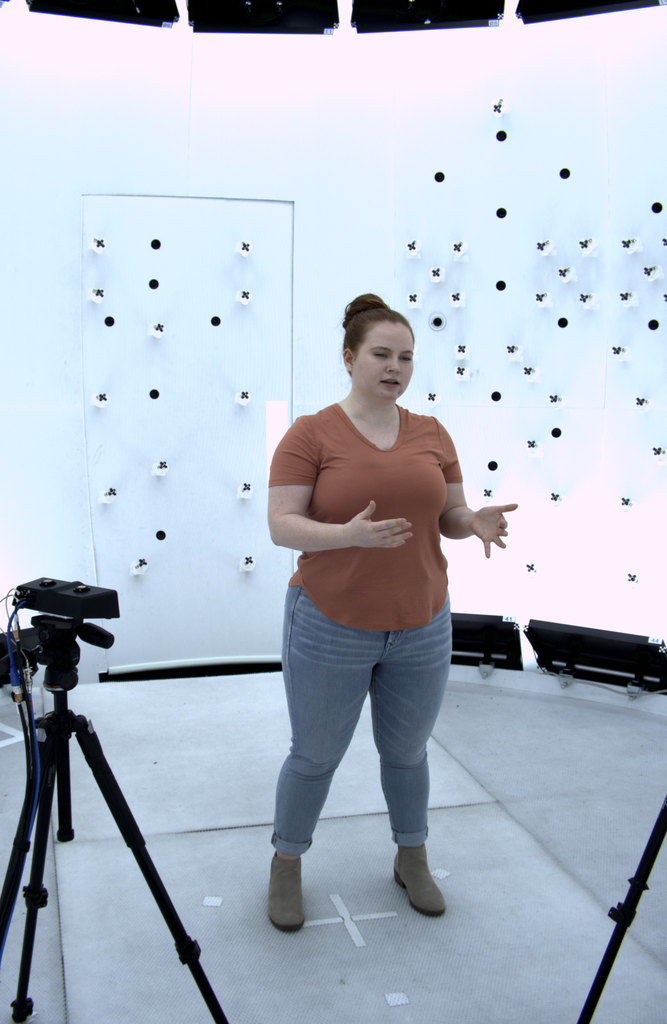}&
\includegraphics[trim={4cm 2cm 2.5cm 8.5cm},clip,width=0.24\textwidth]{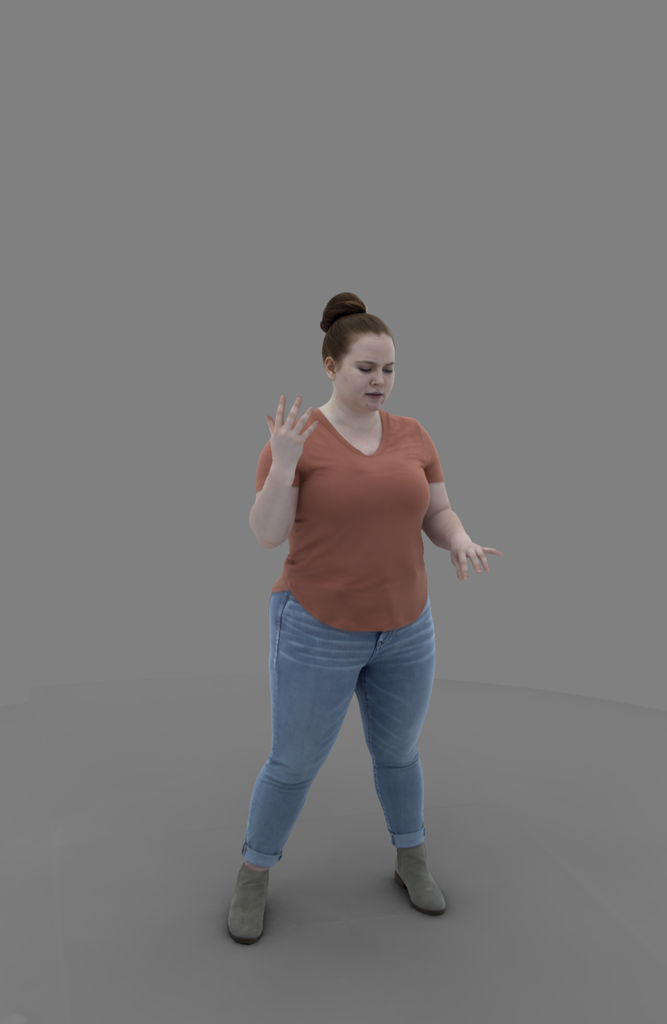}&
\includegraphics[trim={4cm 2cm 2.5cm 8.5cm},clip,width=0.24\textwidth]{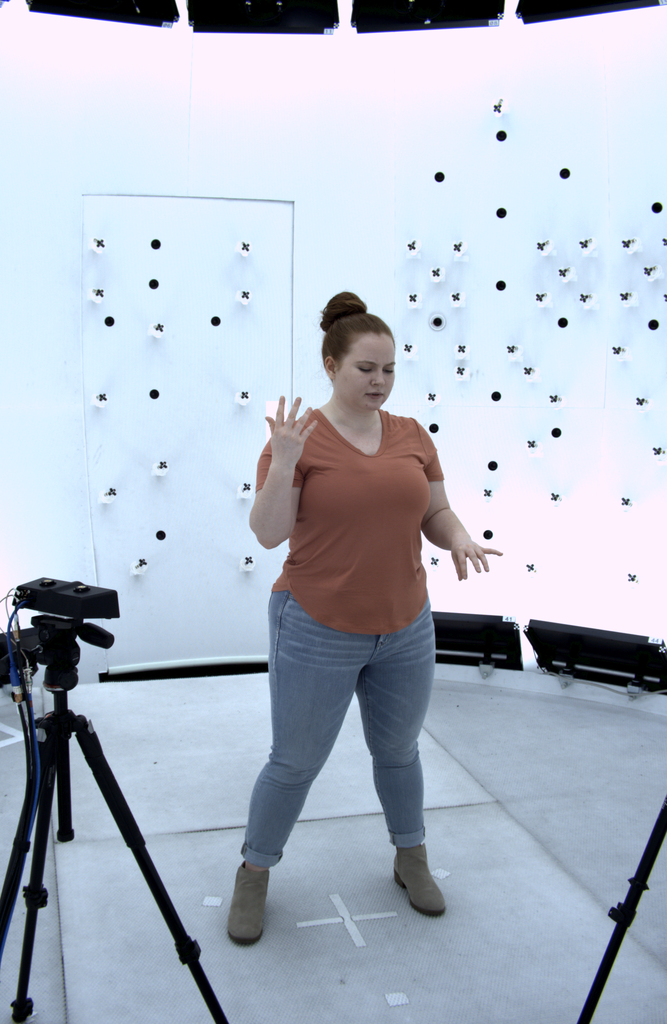}\\
\includegraphics[trim={2.5cm 2cm 2.5cm 5cm},clip,width=0.24\textwidth]{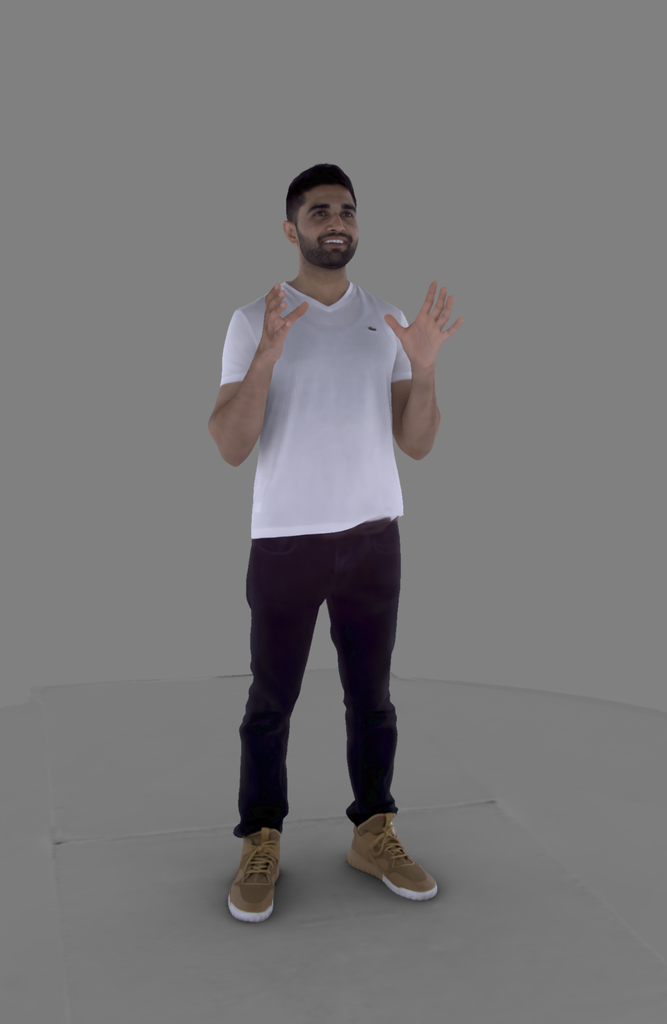}&
\includegraphics[trim={2.5cm 2cm 2.5cm 5cm},clip,width=0.24\textwidth]{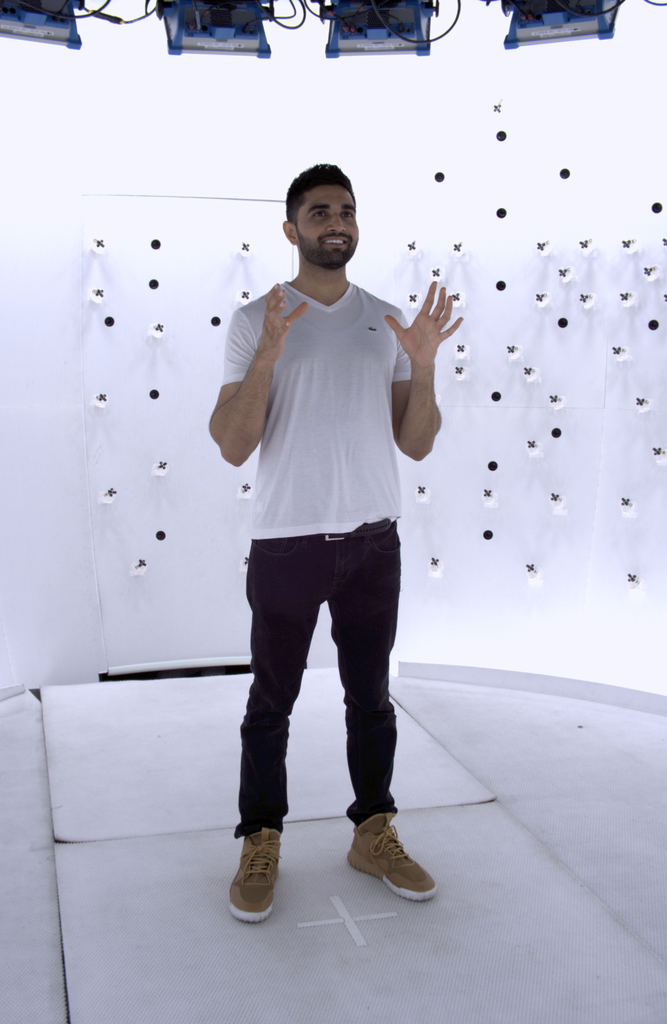}&
\includegraphics[trim={2.5cm 2cm 2.5cm 5cm},clip,width=0.24\textwidth]{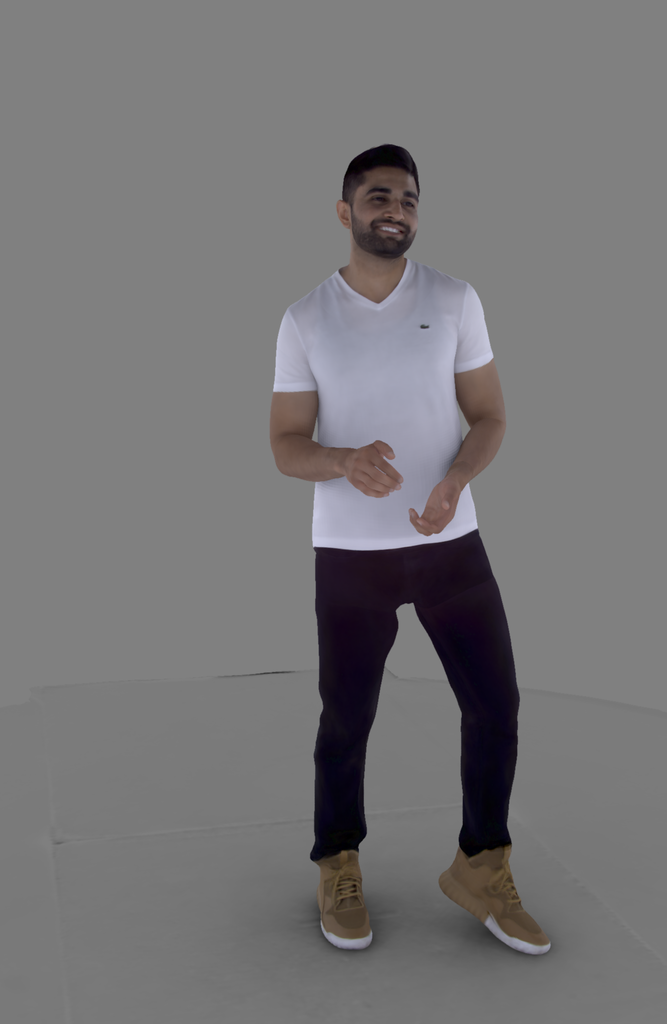}&
\includegraphics[trim={2.5cm 2cm 2.5cm 5cm},clip,width=0.24\textwidth]{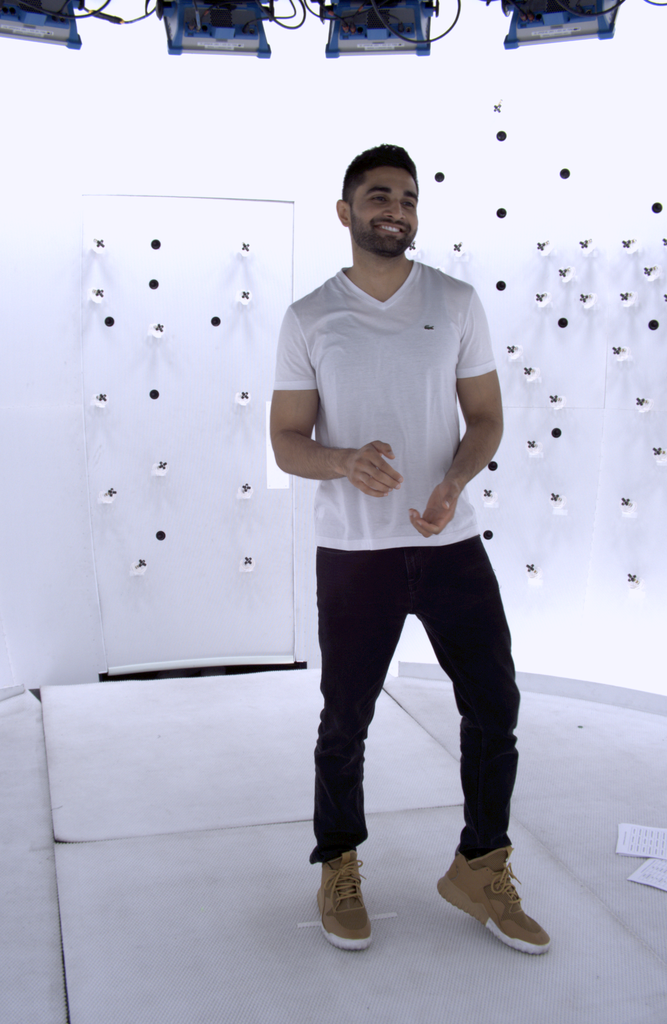}\\
reconstruction & captured image & reconstruction & captured image \\
\end{tabular}
\caption{\textbf{Qualitative Evaluation: Reconstruction Quality.} 
We demonstrate the ability of our model to produce high-quality
reconstructions given full inputs (body pose $\btheta$, facial keypoints $\bbf$ and latent codes $\bz$). 
Note that none of these frames were observed during training.}
\label{fig:results:reconstruction-identities}
\end{figure*}

\begin{figure*}[htp!]
    \centering
    \begin{tabular}{c@{}c@{}c@{}c}
\includegraphics[trim={2.5cm 1.5cm 3cm 6.5cm},clip,width=0.24\textwidth]{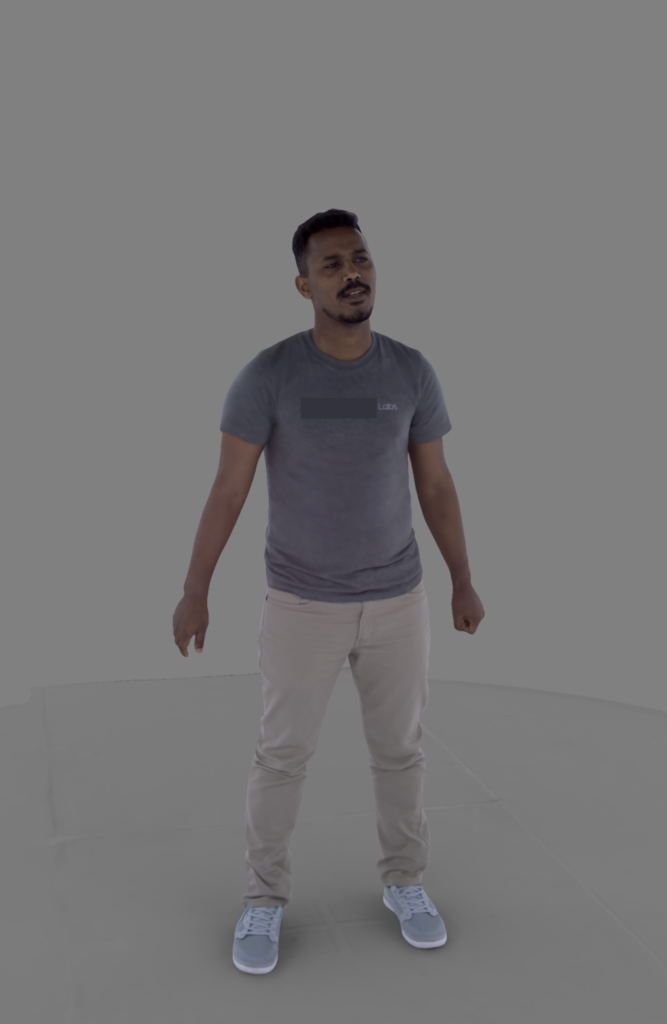}&
\includegraphics[trim={2.5cm 1.5cm 3cm 6.5cm},clip,width=0.24\textwidth]{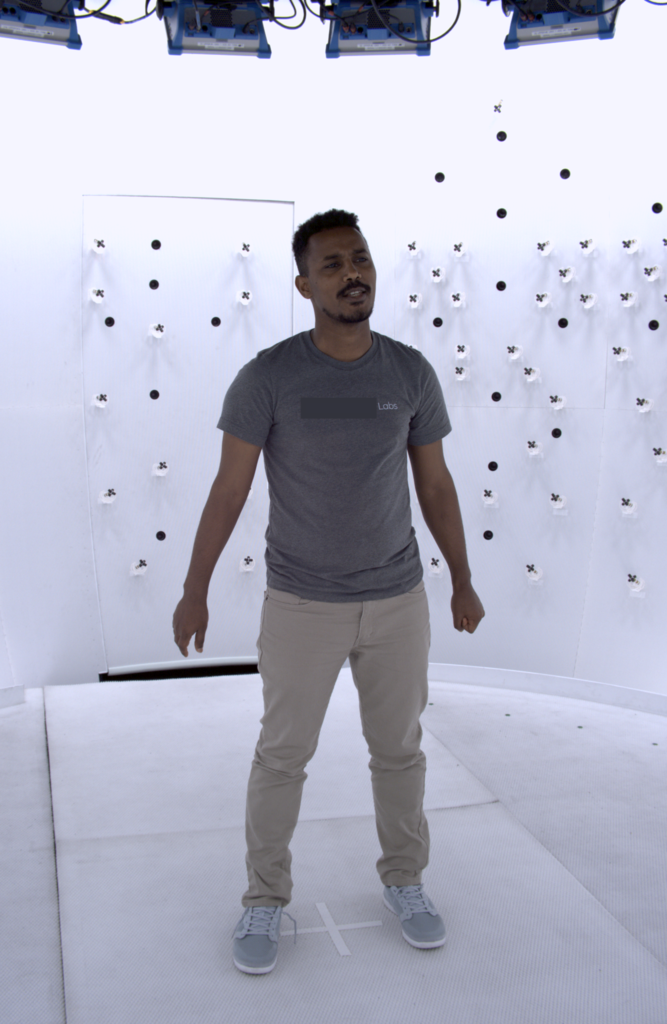}&
\includegraphics[trim={2.5cm 1.5cm 3cm 6.5cm},clip,width=0.24\textwidth]{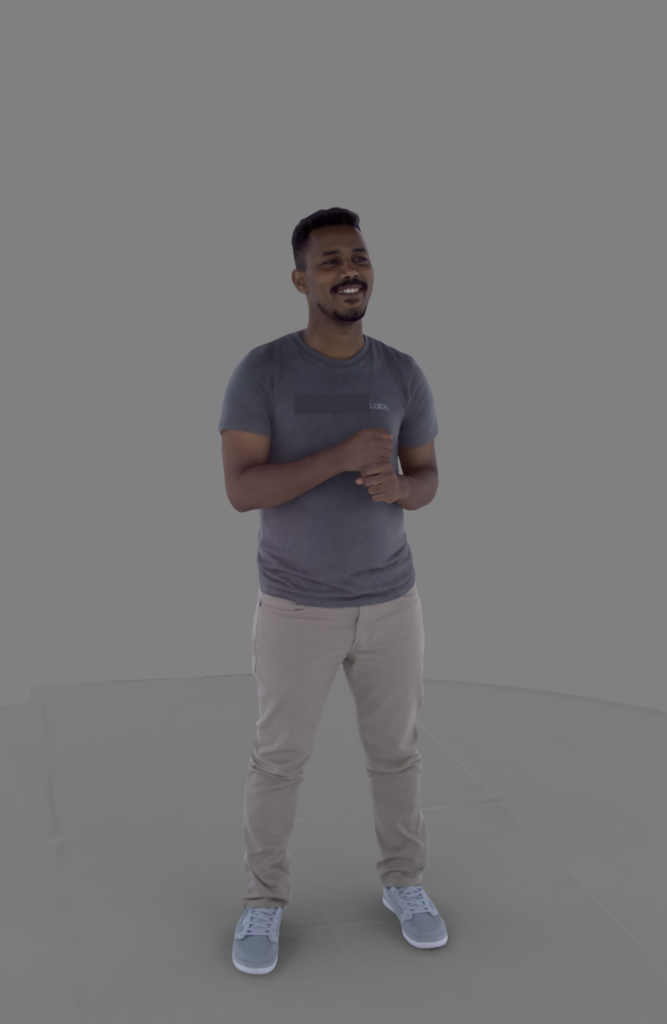}&
\includegraphics[trim={2.5cm 1.5cm 3cm 6.5cm},clip,width=0.24\textwidth]{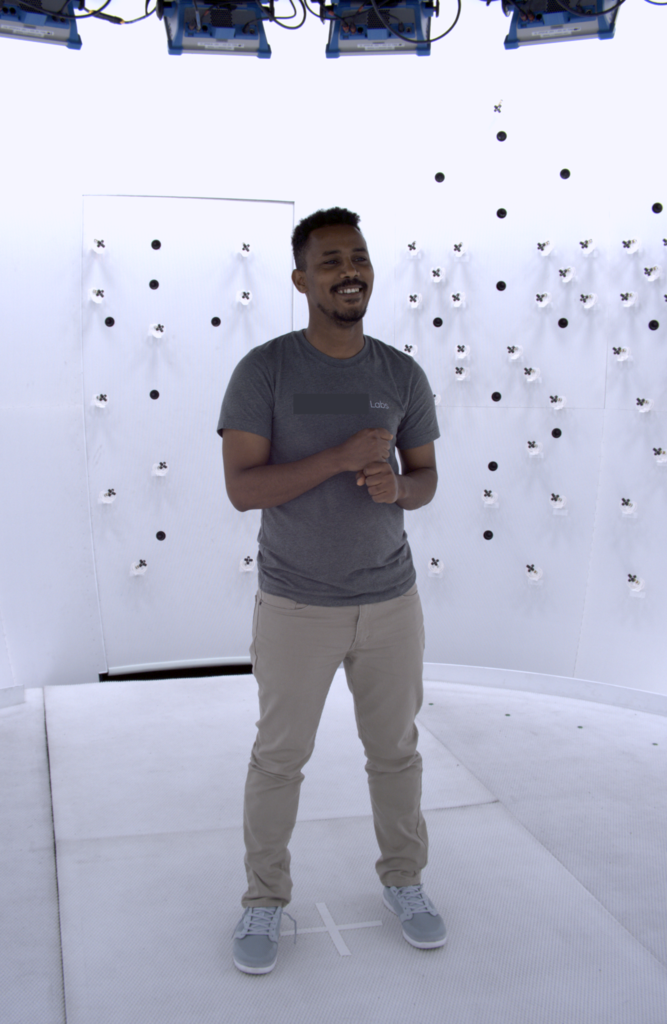}\\
\includegraphics[trim={4cm 2cm 2.5cm 9cm},clip,width=0.24\textwidth]{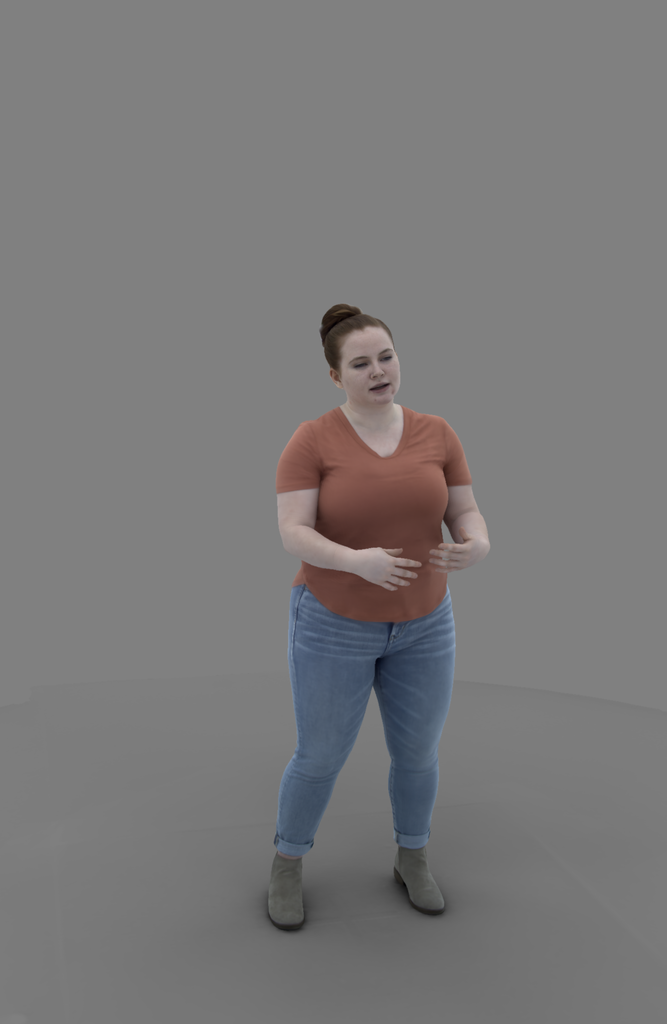}&
\includegraphics[trim={4cm 2cm 2.5cm 9cm},clip,width=0.24\textwidth]{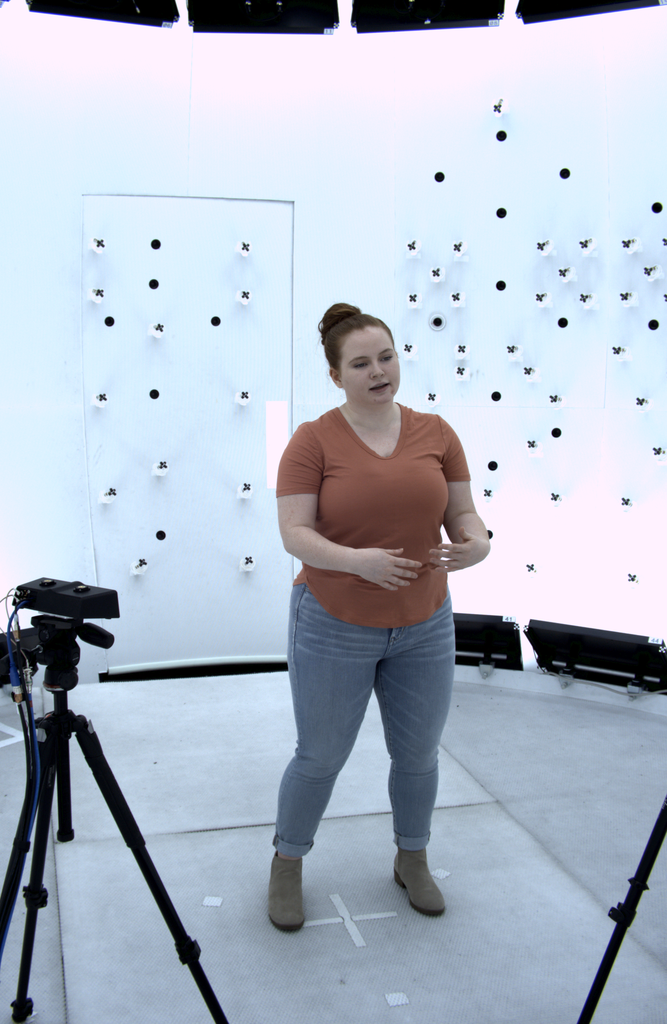}&
\includegraphics[trim={4cm 2cm 2.5cm 9cm},clip,width=0.24\textwidth]{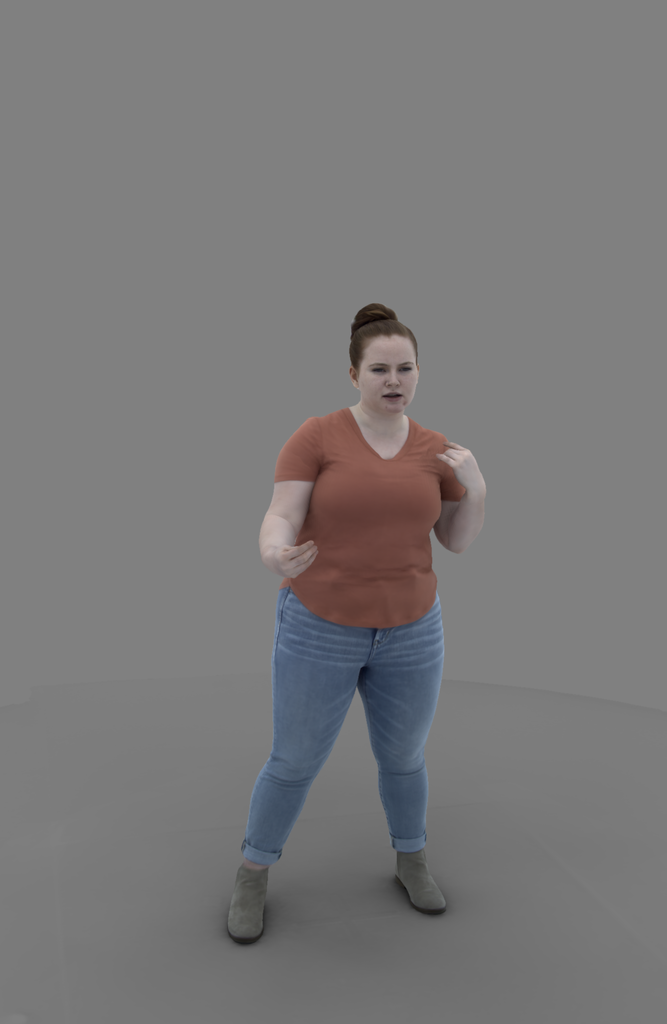}&
\includegraphics[trim={4cm 2cm 2.5cm 9cm},clip,width=0.24\textwidth]{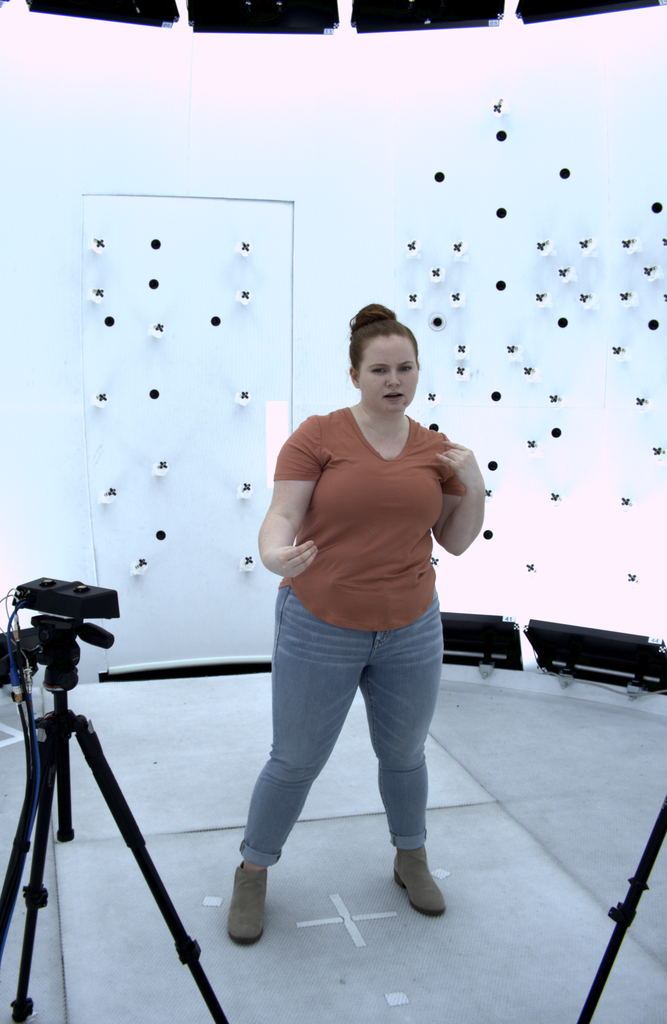}\\
\includegraphics[trim={4cm 3cm 3cm 5.5cm},clip,width=0.24\textwidth]{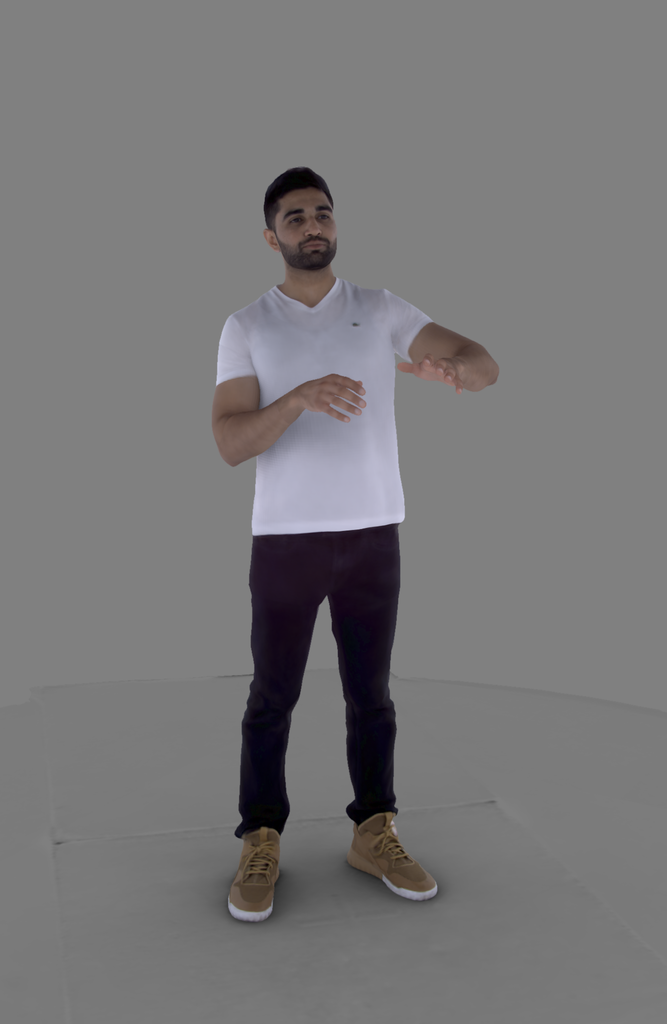}&
\includegraphics[trim={4cm 3cm 3cm 5.5cm},clip,width=0.24\textwidth]{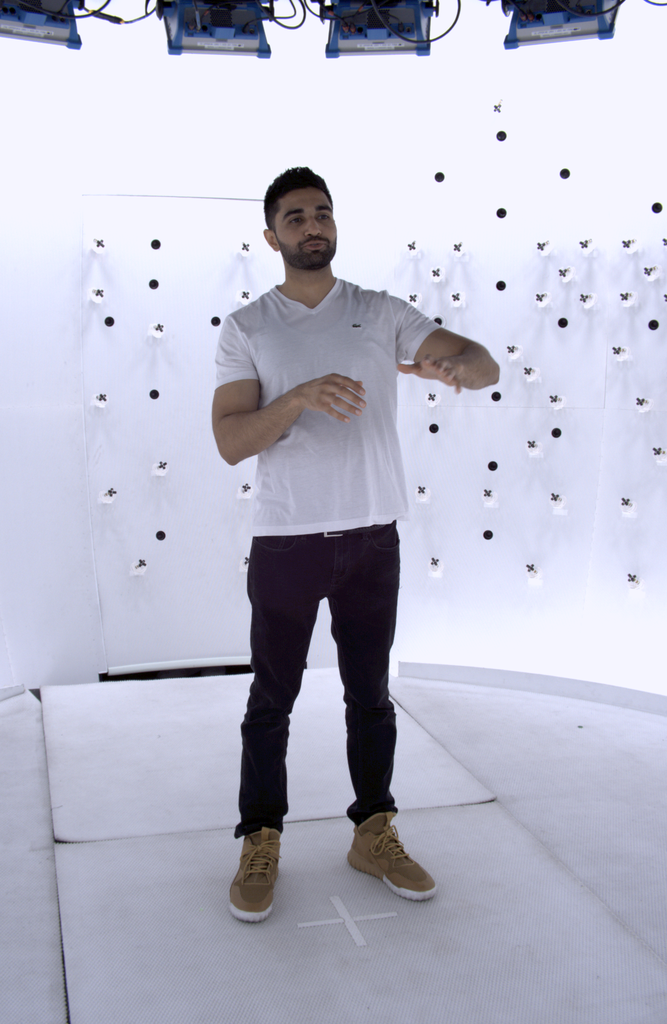}&
\includegraphics[trim={4cm 3cm 3cm 5.5cm},clip,width=0.24\textwidth]{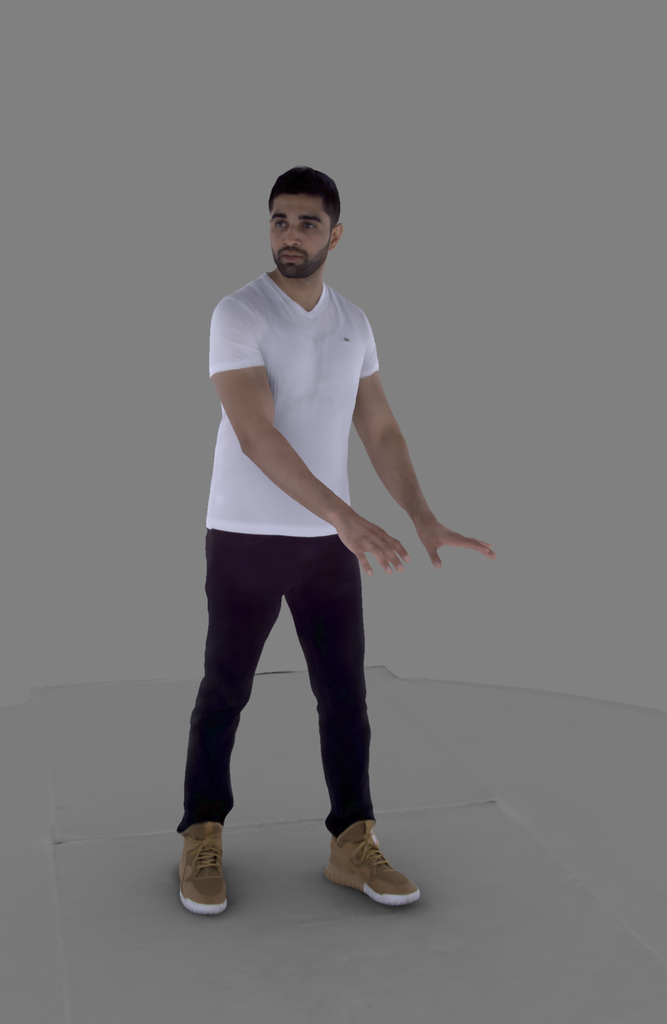}&
\includegraphics[trim={4cm 3cm 3cm 5.5cm},clip,width=0.24\textwidth]{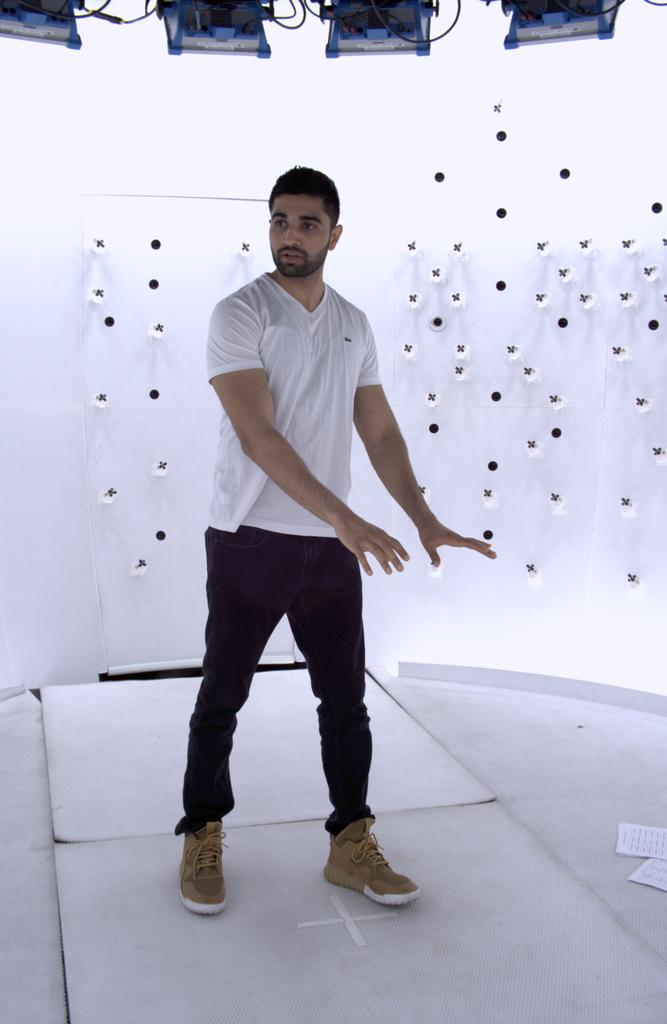}\\
OURS & captured image & OURS & captured image \\
\end{tabular}    
    \caption{\textbf{Qualitative Results: Driving}. For all identities the model
    is conditioned on pose $\btheta$, facial keypoints $\bbf$, and an imputed
    latent code $\bz$. We use a naive imputation strategy with a constant latent code. 
    } 
    \label{fig:results:driving-identities}
\end{figure*}

In this section, we provide an experimental evaluation of our approach
for creating realistic full-body human avatars.
We first shortly describe our capture setup and the data.
We then provide qualitative results on multiple identities on reconstruction
and driving tasks.
Finally, we provide a quantitative evaluation and an ablation
study on different components of our model.

\subsection{Data}

Our capture setup is a multi-camera dome-shaped rig with 140 synchronized cameras, each capable of producing 4096x2668 images.
We collect data from three subjects, where for each subject, we capture sequences of 50-70k frames in length, 
which include a range of motion and natural conversation sequences.
Out of those frames roughly 5000 were left out for testing purposes.
For one of the subjects, we also collected two additional testing sequences:
one where the individual is wearing different clothing,
and another where they are wearing a VR headset. 
In both cases, the full state of the model built from the subject's original capture is not observable. 
In the first, the specific state of clothing is not transferable due to differences in attire between the two captures. 
In the second, parts of the face are occluded by the VR headset. 

To train our full body models, 
we need to obtain skeletal poses as well as registered
surface meshes 
for every frame of the multi-view captures.
For this, we employ a traditional computer vision pipeline 
that is capable of generating reasonable estimates for both.
Specifically,
we follow a two-step approach: first, we run multi-view 3d 
reconstruction~\cite{Galliani15}, keypoint detection and triangulation~\cite{tan2020efficientdet} and foreground body segmentation~\cite{kirillov2020pointrend}. 
Second, we perform skeletal pose estimation using a personalized body rig and LBS by matching the various features computed in the first step as well as pose priors over joint angles~\cite{gall2009motion}. 
We further refine these estimates by running a deformable ICP algorithm on top of the tracked meshes so as to fit the 3D scan details better with surface Laplacian regularization~\cite{Sorkine04sgp}. One example from the various stage of the pipeline can be seen in Figure~\ref{fig:data-processing}.
Processing the entire dataset for a single subject takes ~14 days with 160 GPUs (20 DGX servers with 8 GPU each). This data processing pipeline is robust enough to process most of the data, and only has challenges in poses with heavy self-occlusions, like heavy hand-cloth interactions. We discard on average roughly 10 percent of frames by filtering out frames where the tracked surface has large error compared with the scans. Notice that even on successfully registered frames, due to limited 3D scan resolution or foreground mask inaccuracies, the registered mesh may exhibit errors in shape or correspondence. We observe that these errors are greatly reduced during model training by using inverse rendering with an image reconstruction loss. One example of such improvement can be seen in Figure~\ref{fig:inverse-rendering}, where the decoded mesh shows greatly improved alignment with respect to the initial tracking mesh in high detail areas such as the fingers and around creases or folds.

\subsection{Qualitative Results}

\noindent \subsubsection{Reconstruction} 
We first evaluate our model's ability to generalize to new poses that are held out from the training set. 
For this we optimize all the model's parameters, including the global pose, skeletal joint angles $\boldsymbol{\theta}$, face parameters $\mathbf{f}$ and latent codes $\mathbf{z}$, to minimize the reconstruction loss over images from the multi-view cameras as well as the tracked mesh. 
Figure~\ref{fig:results:reconstruction-identities} shows
the reconstruction results for three different identities and
different combinations of poses and facial expressions, 
along with the ground truth capture images as a reference.
The results demonstrate good reconstruction, capable of capturing subtle details of expression and pose. 

\noindent \subsubsection{Driving} 
\label{sec:experiment_driving}
Our main goal is to build drivable models, which produce
realistically looking virtual humans given 
real
driving signals.
In this section, we thus evaluate the quality of our model 
on the task of \textit{driving}, 
where only partial information about the full model's state is available via the driving signals. 
The latent codes are not provided and have to be imputed.

\begin{figure*}[ht!]
    \centering
    \includegraphics[width=\textwidth]{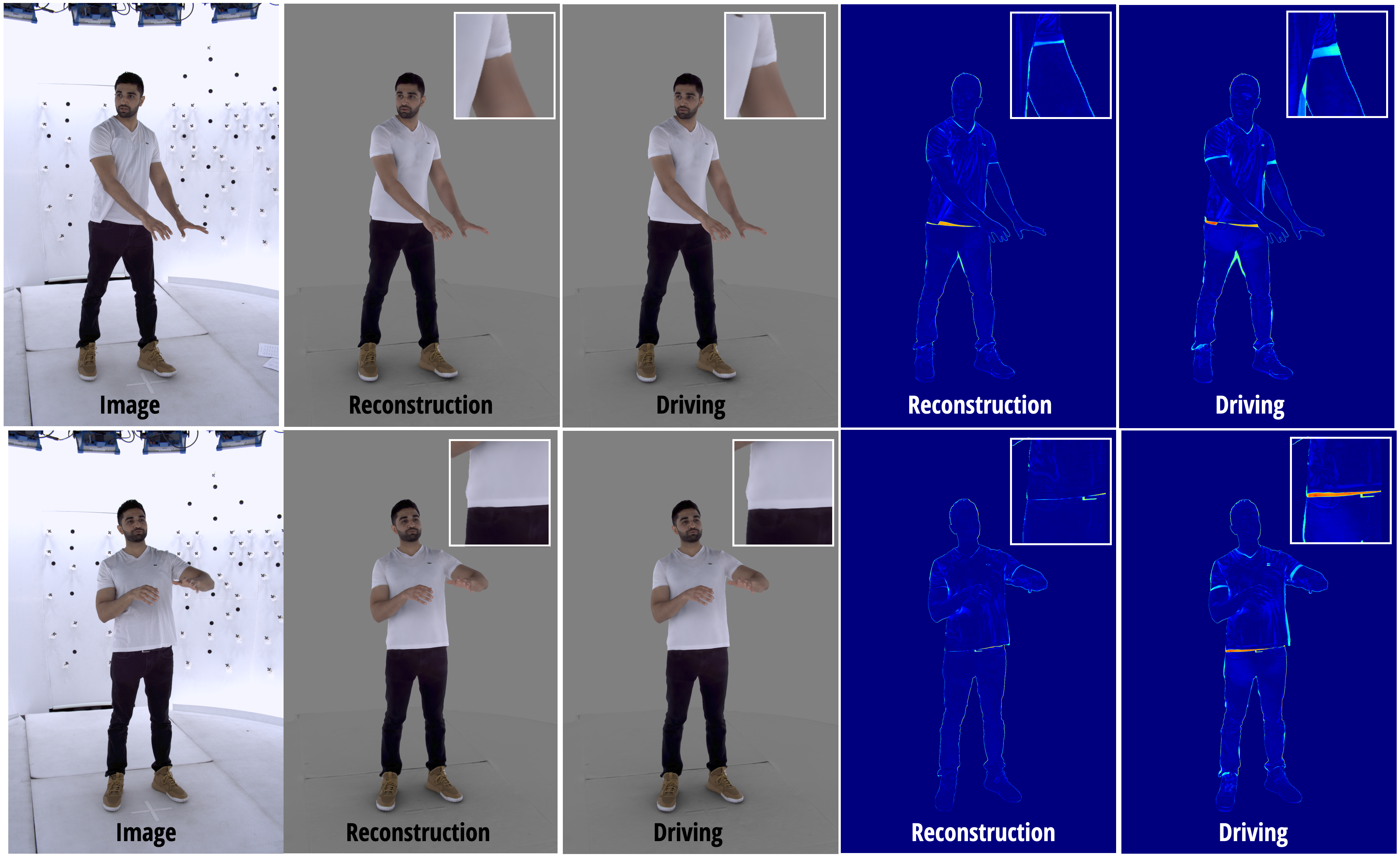}
    \caption{\textbf{Information Deficiency}. Driving by setting the latent code to zero produces plausible results but does not match the data where the driving signal is deficient, for example to model specific clothing states. } 
    \label{fig:results:information-deficiency}
\end{figure*}

\begin{figure*}[ht!]
    \centering
    \begin{tabular}{c@{}c@{}c@{}c}
\includegraphics[trim={4.5cm 1.5cm 3cm 8.5cm},clip,width=0.24\textwidth]{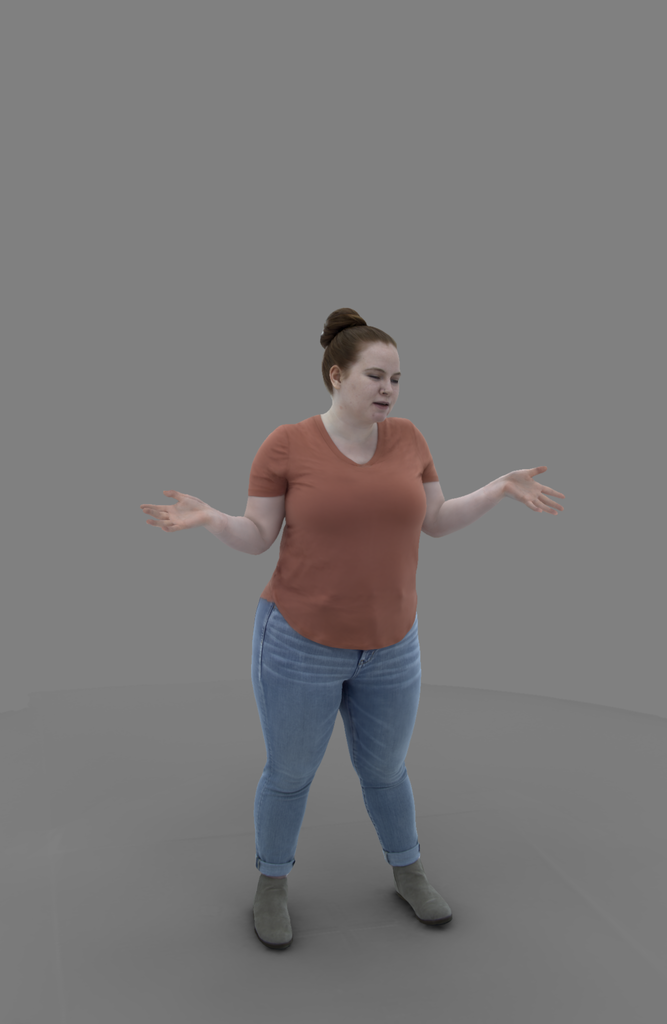}&
\includegraphics[trim={4.5cm 1.5cm 3cm 8.5cm},clip,width=0.24\textwidth]{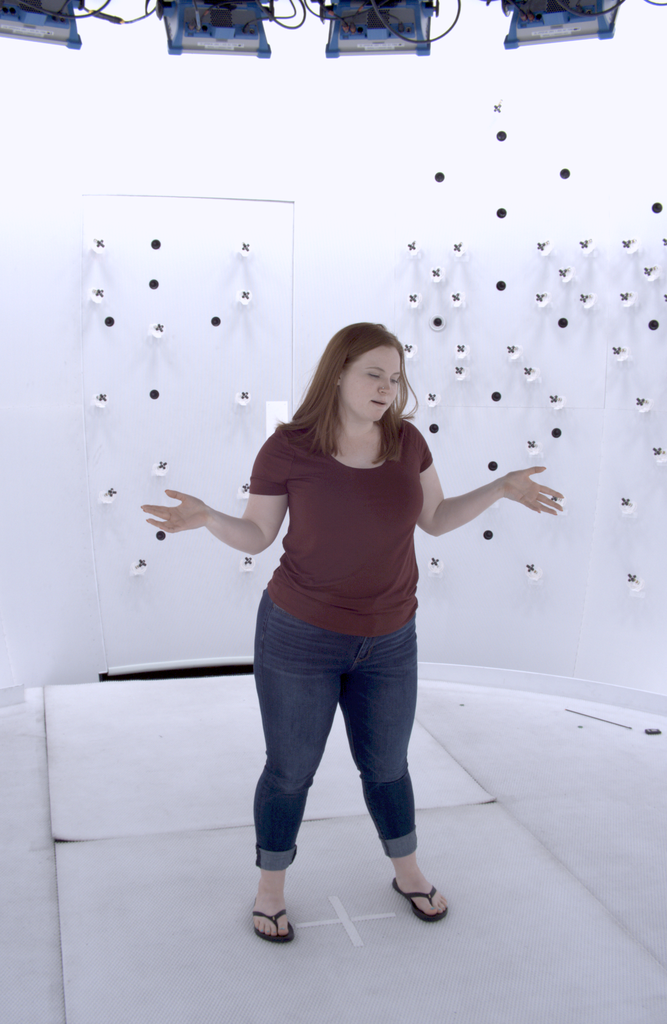}&
\includegraphics[trim={4.5cm 1.5cm 3cm 8.5cm},clip,width=0.24\textwidth]{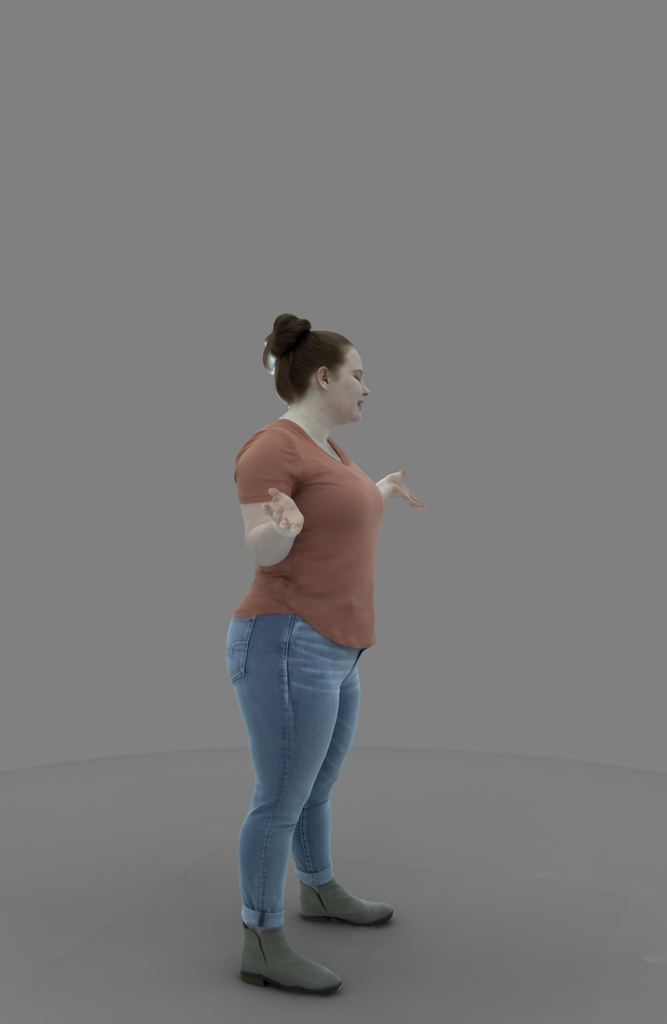}&
\includegraphics[trim={4.5cm 1.5cm 3cm 8.5cm},clip,width=0.24\textwidth]{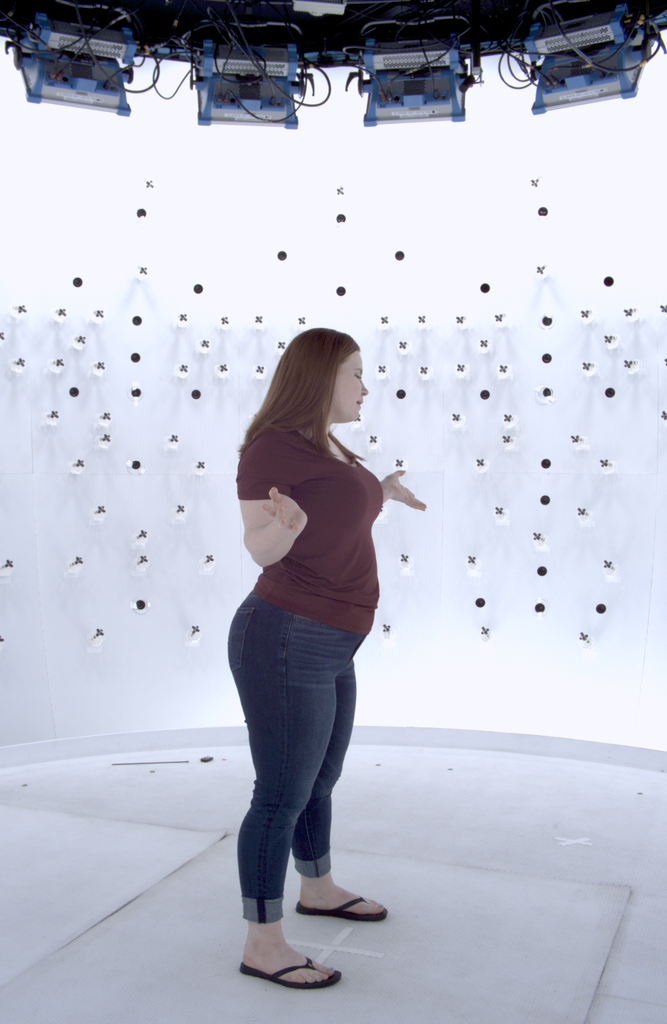}\\
\end{tabular}    
    \caption{\textbf{Qualitative Evaluation: Driving in Different Clothing}. 
    The model is conditioned
    on pose $\btheta$, facial keypoints $\bbf$, and an imputed latent code $\bz$.
    We use a naive imputation strategy with constant latent code. } 
    \label{fig:results:driving-bodies-pilot}
\end{figure*}

\begin{figure*}[ht!]
    \centering
    \begin{tabular}{c@{}c@{}c@{}c}
    \includegraphics[trim={4.5cm 1.5cm 3cm 8.5cm},clip,width=0.24\textwidth]{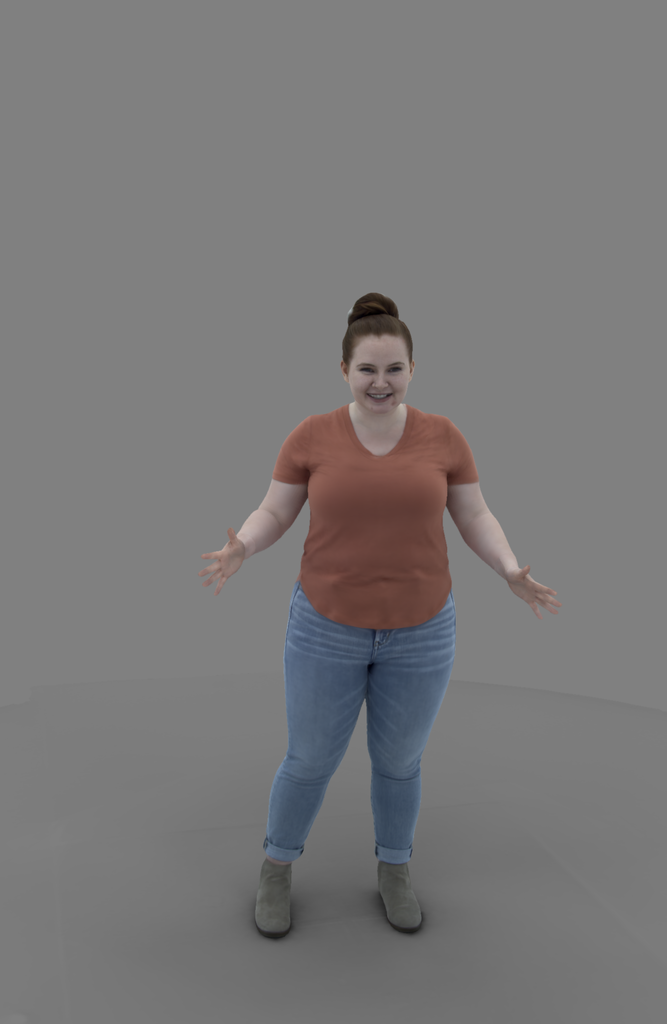}&
    \includegraphics[trim={4.5cm 1.5cm 3cm 8.5cm},clip,width=0.24\textwidth]{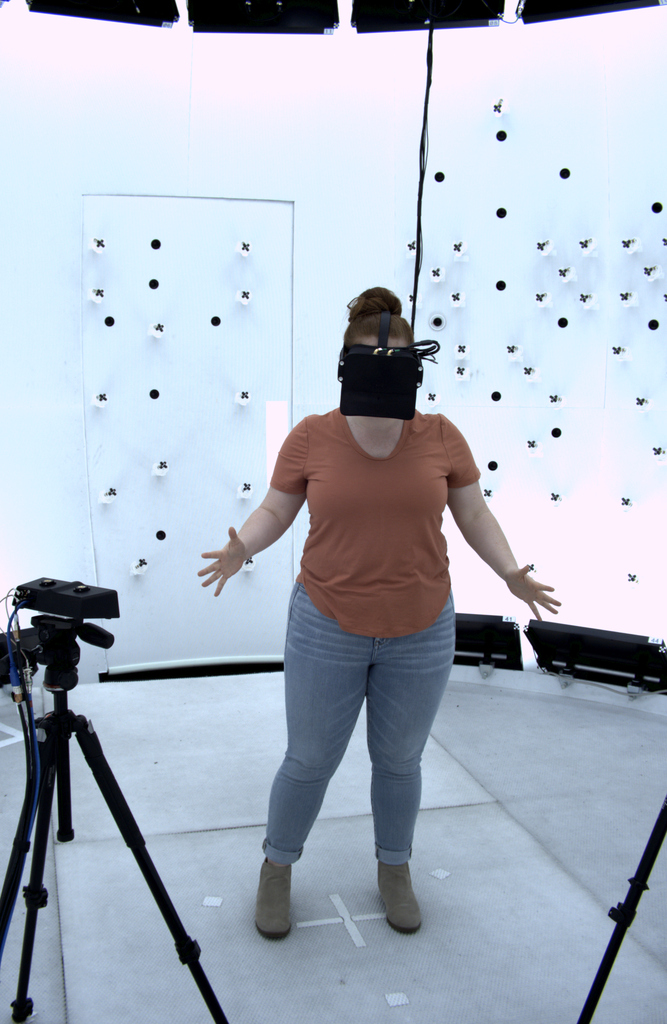}&
    \includegraphics[trim={4.5cm 0.5cm 3cm 9.5cm},clip,width=0.24\textwidth]{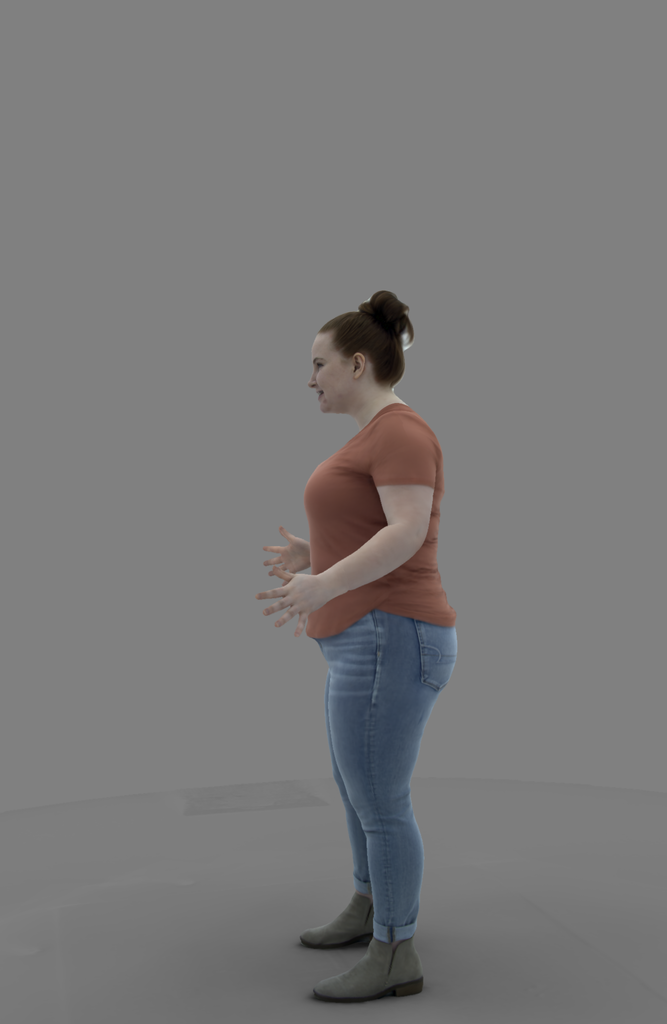}&
    \includegraphics[trim={4.5cm 0.5cm 3cm 9.5cm},clip,width=0.24\textwidth]{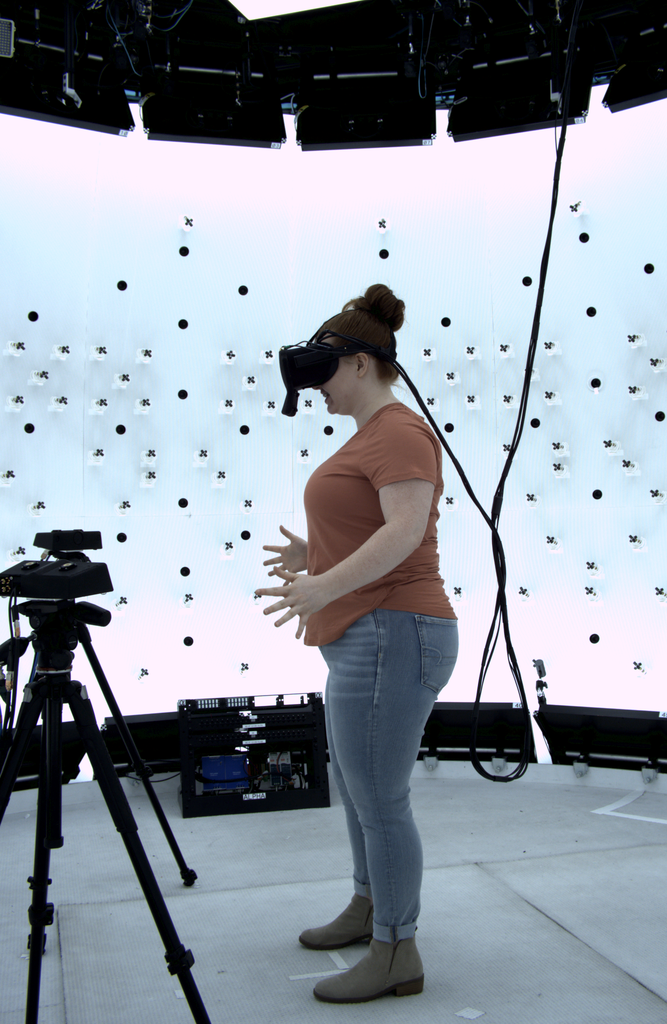}\\
    \end{tabular}    
    \caption{\textbf{Qualitative Evaluation: Driving with a Headset}. The model is conditioned
    on pose $\btheta$, facial expression codes $\bbf$, and an imputed latent code $\bz$.} 
    \label{fig:results:driving-bodies-headset}
\end{figure*}

\noindent\paragraph{Matched Setting}

First we would like to evaluate how much information is really missing from a full body model that is not accounted for by skeletal joint angles and facial expressions. 
In Figure~\ref{fig:results:driving-identities} we show driving results where only the pose $\boldsymbol{\theta}$ and facial keypoints $\mathbf{f}$ are presumed to be available, and we simply set the latent code to zero; a maximum likelihood imputation. 
A video sequence of this driving can be found in the supplemental video. 
The first point to note is that the pose and facial expressions appear to be matched well between the avatar and the ground truth images, demonstrating the efficacy of the disentanglement scheme described in \S\ref{sec:method}. 
The second point to notice, is that the reconstruction attained by the driven model is of less quality than the reconstructions, with errors concentrated around clothing; areas where we expect joint and 
facial keypoints to have limited disambiguation capabilities. 
A comparison between driving and reconstruction can be found in Figure~\ref{fig:results:information-deficiency}.

\begin{figure*}[ht!]
    \centering
    \begin{tabular}{c@{}c@{}c@{}c}
\includegraphics[trim={4.5cm 2.5cm 3cm 10cm},clip,width=0.24\textwidth]{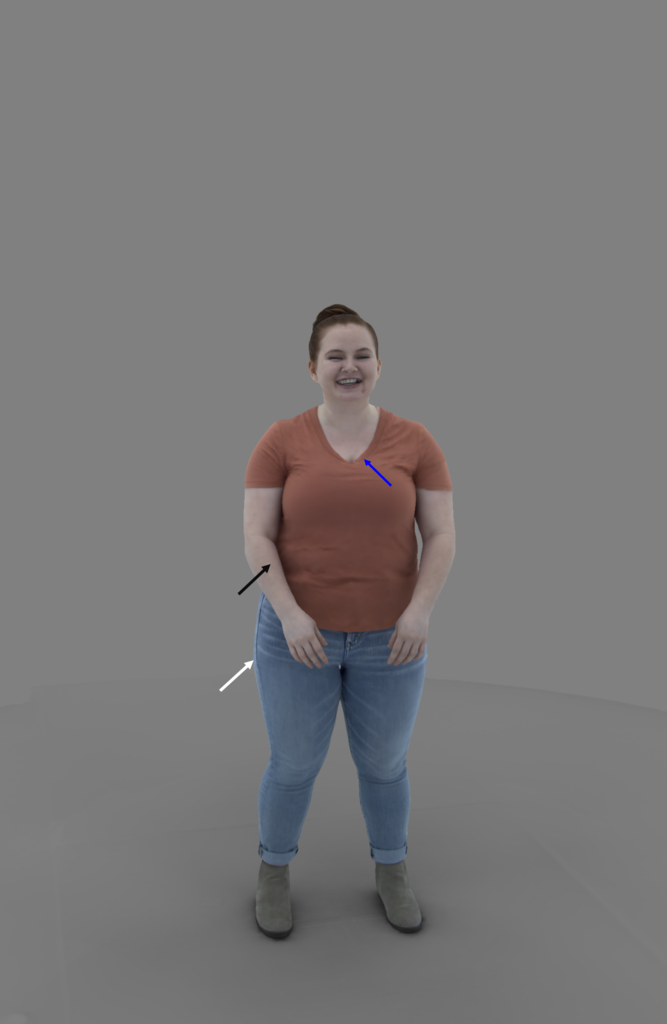}&
\includegraphics[trim={4.5cm 2.5cm 3cm 10cm},clip,width=0.24\textwidth]{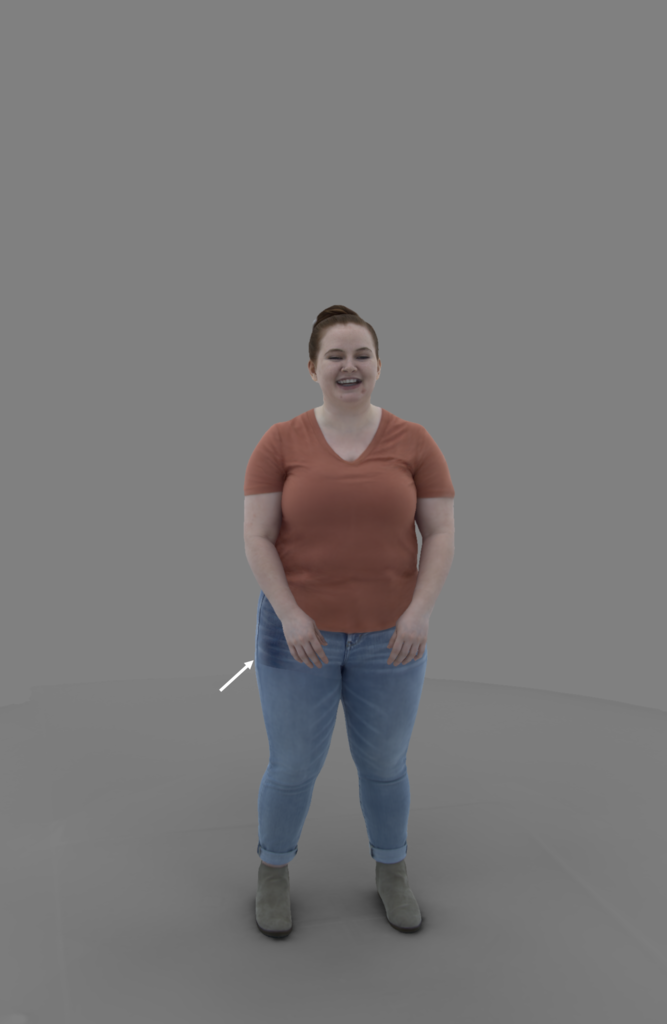}&
\includegraphics[trim={4.5cm 2.5cm 3cm 10cm},clip,width=0.24\textwidth]{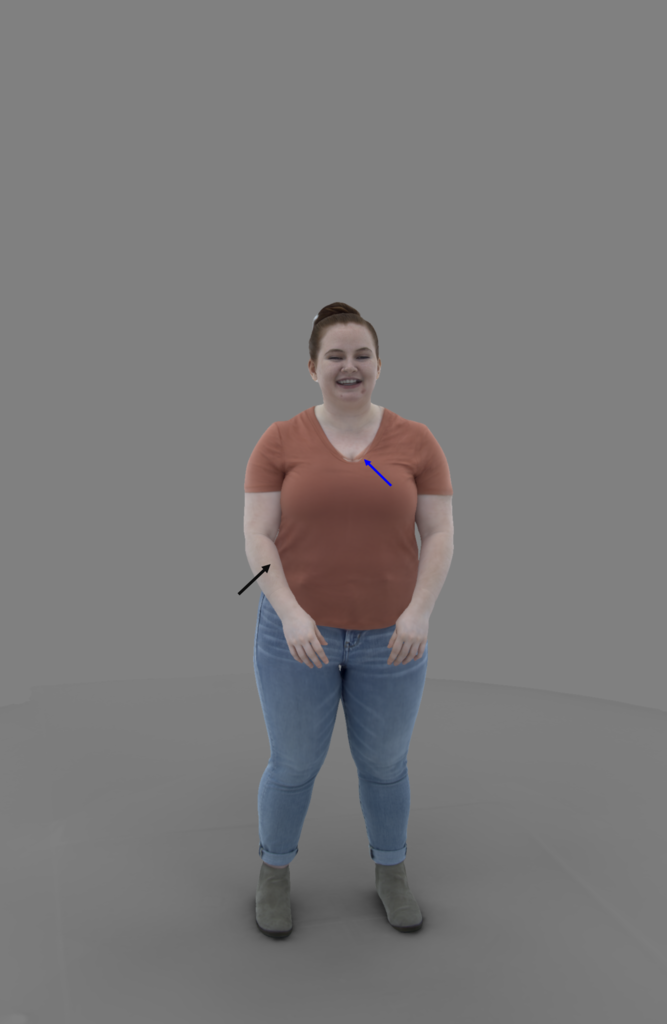}&
\includegraphics[trim={4.5cm 2.5cm 3cm 10cm},clip,width=0.24\textwidth]{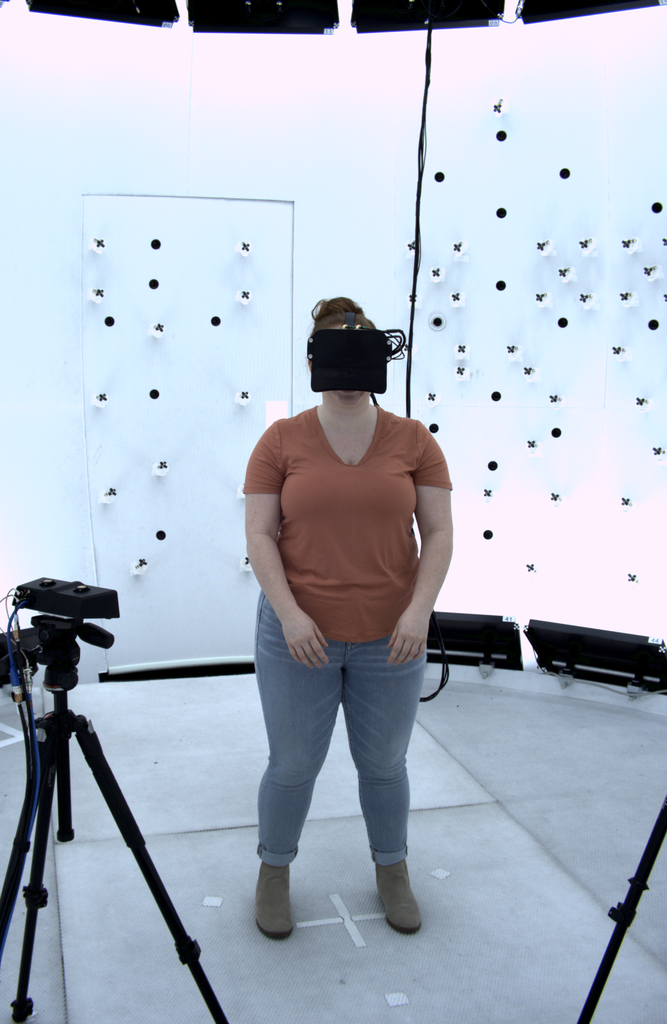}\\
\includegraphics[trim={5.5cm 2.5cm 3cm 11cm},clip,width=0.24\textwidth]{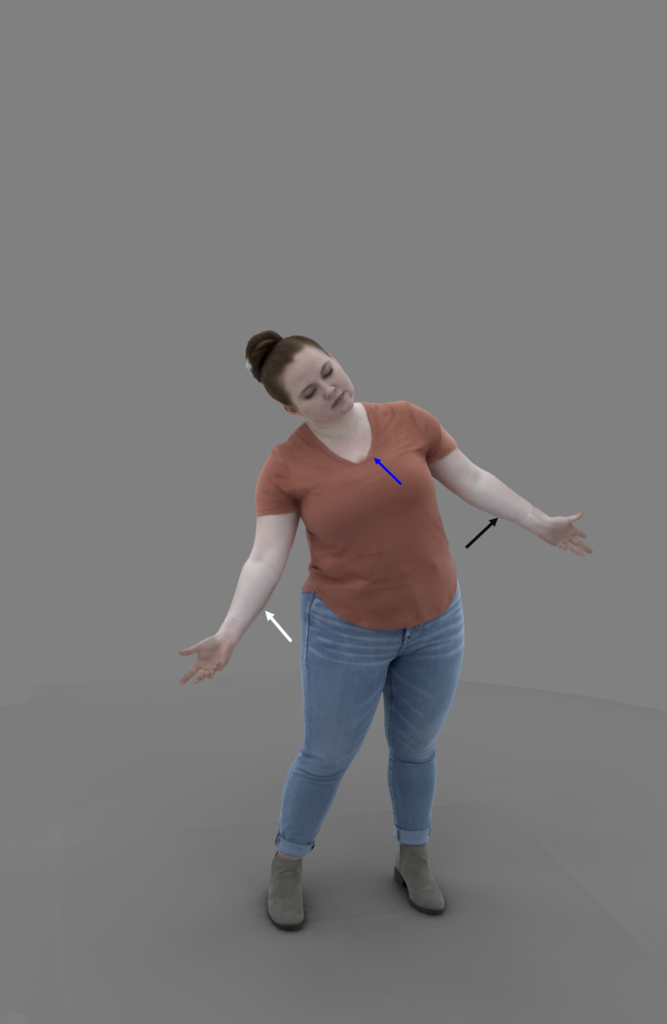}&
\includegraphics[trim={5.5cm 2.5cm 3cm 11cm},clip,width=0.24\textwidth]{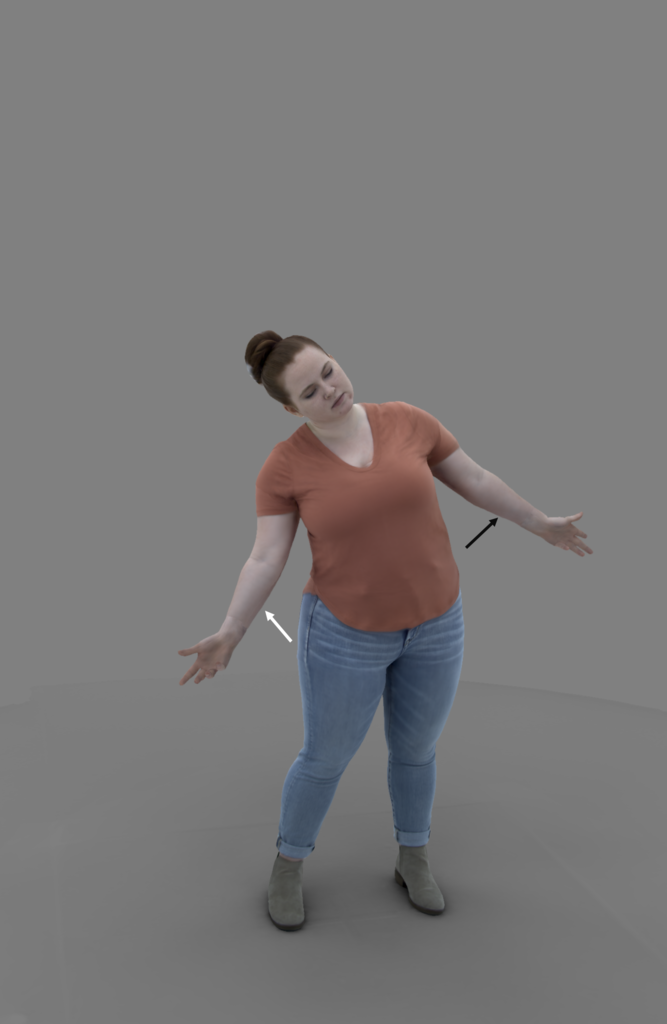}&
\includegraphics[trim={5.5cm 2.5cm 3cm 11cm},clip,width=0.24\textwidth]{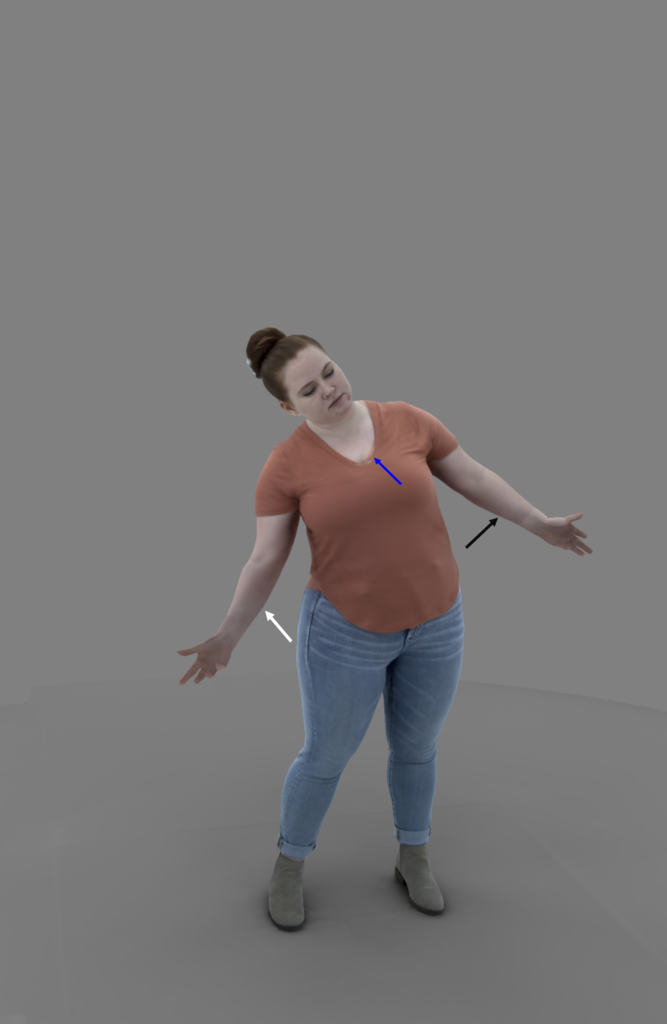}&
\includegraphics[trim={5.5cm 2.5cm 3cm 11cm},clip,width=0.24\textwidth]{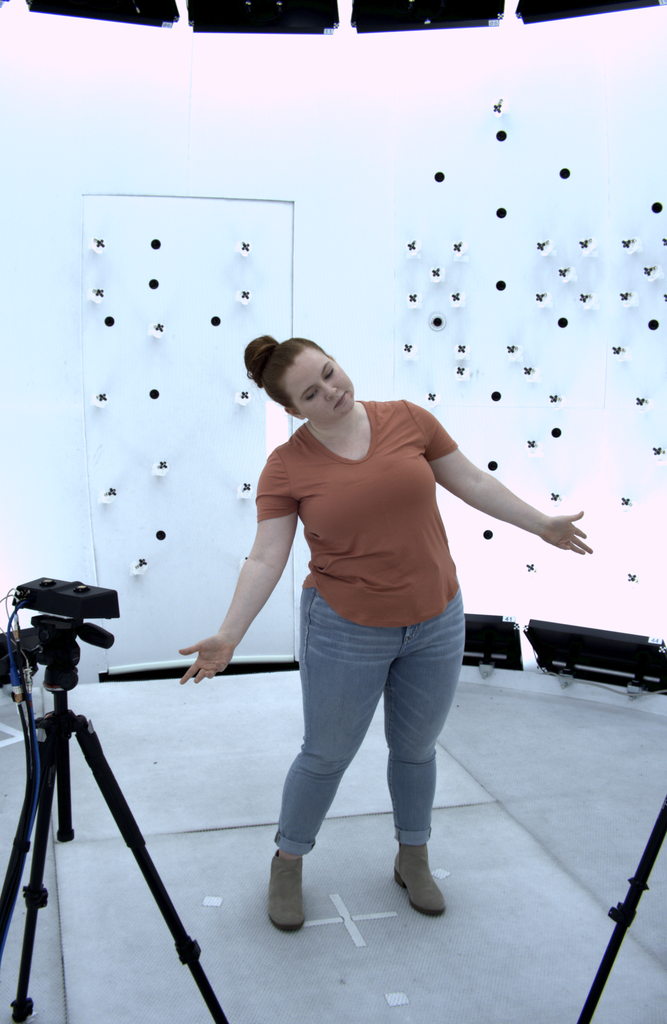}\\
OURS & pose+face & pose+face+latent & capture image 
\end{tabular}    
    \caption{\textbf{Qualitative Comparison}. We use a naive imputation strategy with a constant latent code. Both baselines struggle at capturing shadows (white and black arrows) and avoiding ghost artefact (blue arrows). } 
    \label{fig:results:driving-comparison}
\end{figure*}

Now that we have established the extent of missing information, we would like to evaluate the effects this has on models that ignore it or model it without disentangling. In Figure~\ref{fig:results:driving-comparison} we provide a comparison
between our method and two baselines: one without the latent space (pose+face)
and one without latent-space disentanglement (pose+face+latent).
Both baselines appear to exhibit a number of visual artifacts, including incorrect shadows, ghosting, and over-smoothing. 
For example, extraneous shadows on the pants in (pose+face) and missing shading in the armpits for (pose+face+latent) in the top row images. Another example is the unnatural wrinkles in the torso for (pose+face) and ghosting artefact in the neck area of the shirt for (pose+face+latent) in the bottom row. 
At a glance, static views of these models without a latent space can look acceptable, but over-fitting artifacts are significantly more noticeable in dynamic sequences. We encourage the reader to view the accompanying supplementary materials, in which we also provide additional comparison to an image-space method~\cite{thies2019deferred}.
%

\noindent\paragraph{Unmatched Setting} In the matched setting above, information deficiency was contrived in order to evaluate its effect on animation. In practice, one could have chosen to solve for the latent code that minimizes the reconstruction error since the capture setting for the model and the performer are identical. However, a far more common scenario is when there is a domain gap between the capture setting during modeling and during driving. In this case, the model can not fully span the appearance in the driving image, and must rely instead on robust features that can be equivalently extracted in both domains to drive the model. We consider two such scenarios for one of our capture subjects. 

In the first, the subject is captured in a different attire; different colored clothing, sandals instead of shoes, and hair out instead of tied in a bun. This mimics the scenario where a new performance for free-viewpoint video needs to be captured at a different time, where the actor's appearance has changed or the old attire is not accessible. Here, the skeletal pose and facial keypoints extracted from the capture studio are used to drive the avatar. Some frames of this capture along, with the animated avatar, are shown in Figure~\ref{fig:results:driving-bodies-pilot}. 

In the second, the subject is driving the model while wearing a VR headset which occludes parts of her face. There is a pair of stereo cameras in the scene as well as on the headset that extract driving signals that comprise the skeletal keypoints and facial expression codes obtained using the approach in~\cite{wei2020VRFace}. This setting is a minimal sensing configuration for enabling VR-based telepresence for two-way interaction. Some frames demonstrating the performance of the system are shown in Figure~\ref{fig:results:driving-bodies-headset}.

In both the attire and headset settings, our model remains faithful to the information content present in the driving signal, while producing plausible imputations of information that is missing by simply setting the latent code in all cases to zero. Further results of these driving results can be found in the supplementary video, where we additionally visualize results that use different methods for latent code synthesis, such as through random sampling and by using a temporal model. By disentangling the latent and driving signals, our method opens up the possibility of applying a variety of different imputation strategies that could be suited to particular applications. A deeper investigation into latent code synthesis is out of the scope of this paper, but it is an interesting direction of future work.

%
%

\subsection{Quantitative Evaluation}

In Table~\ref{tab:reconstruction}, we report numerical results in terms
of image error with respect to capture images.
Because our main goal is to evaluate how \textit{drivable} the models
are, all models are conditioned on the same set of driving signals 
(i.e. facial expression codes $\bbf$ and skeletal poses $\btheta$),
and all the models that have a latent space are all conditioned on the 
same constant latent code $\bz = \mathbf{0}$.
On the training set, the model without a latent space (i.e. pose+face) 
achieves the lowest error, albeit at the cost of severe 
overfitting, as it suffers a very significant performance drop
on the test set.
Among the variations of our methods, 
the lowest error is obtained 
by a version without compression mechanism, which we also attribute 
to overfitting.
%
%
The baseline with a latent space (pose+face+latent) 
performs worse on both train and test - which 
we attribute to the fact that it does not employ any 
disentanglement-promoting mechanism.
Similarly, out of all the versions of our approach, the model without
an explicit disentanglement performs slightly worse on both.

On the test set, our method has a clear advantage over the model 
without latent space disentanglement (pose+face+latent).
Performance of the model without the shadow branch indicates that
it is useful for generalization; the model without the shadow
branch is able to fit the training data well, but suffers significant degradation on the test set. 
Similarly, the version of our model without spatially localized embeddings
has significantly lower training error, but performs significantly
worse on the test set.
We can also infer that the version without disentangling is
prone to poor generalization; the model is able to fit the training 
data but leads to the worst performance amongst all variants of our model.


\begin{table}[ht!]
    \centering
    \caption{\textbf{Quantitative Evaluation.} Image reconstruction error 
    for driving on training data (train) and testing (unseen) samples.
    The baseline without a latent code (pose+face) is prone
    to learning chance correlations and leads to severe overfitting, 
    the baseline with a latent code (pose+face+latent) without
    disentanglement mechanisms has poor drivability. }
    \label{tab:reconstruction}    
    \begin{tabular}{|l|c|c|}
    \hline
    method &  train & test \\\hline    
    OURS & 11.578 & \textbf{15.546} \\    
    pose+face & \textbf{7.647} & 17.509 \\
    pose+face+latent & 14.861 & 19.885 \\\hline
    OURS (no disent.) & 11.926 & 16.364 \\
    OURS (no spat.local.) & 9.424 & 16.461 \\
    OURS (no shadow) & 10.872 & 16.792\\ \hline
\end{tabular}

\end{table}

\section{Limitations and Future Work}

Our model relies on tracked meshes for supervision and is thus limited by the tracking quality. 
As tracking extremely loose clothing with topological 
changes is still an area of active research, it would be challenging to
apply our model to those complex scenarios. 
Moreover, our network architecture relies heavily on the assumption of fixed topology, 
as it operates on UV-based mesh representation, which limits the 
applicable scenarios, and may be tackled by instead relying on implicit 
surface representations~\cite{park2019deepsdf, remelli2020meshsdf}.
UV-based representations can also be prone to seam artifacts 
(noticable e.g. in Figure~\ref{fig:results:driving-bodies-headset}), especially when combined with 2D 
convolutions~\cite{groueix2018papier}. This issue can be addressed by using 
mesh convolutions~\cite{bronstein2017geometric,YI2020}, albeit at a higher computational cost.

Although latent codes do capture clothing variations and some of the 
high-frequency deformations, the limited capacity of the latent space 
can lead to loss of details, which could be the reason that reconstructions in 
Figure~\ref{fig:results:reconstruction-identities} do not capture
all of the clothing deformations.
One potential future direction to tackle this would be to apply hierarchical 
generative models~\cite{vahdat2020NVAE,Bagautdinov2018}, which tend to have better
representational power.

We employed a naive strategy for imputing missing information by setting the 
latent codes to a constant value for every frame, which means that the state of
clothing will be fixed during animation (of course, pose-dependent deformations
can still be captured).
An interesting direction of future work is to investigate more advanced approaches 
that can be tailored to a specific application. 
For example, employing temporal models or style-dependent generation~\cite{ginosar2019gestures, alexanderson2020speech2gesture} may enable more semantically meaningful imputations. 
At the same time, methods for visualizing the space of uncertainty may be relevant 
for applications that rely on the authenticity of the animation, such as telepresence. 

Our data collection and experimental evaluation are primarily focused on natural
conversations, and thus it is not fully clear if our pipeline generalizes to particularly challenging 
poses which are far apart from the training distribution.
In the supplementary, we provide an interactive tool for t-SNE-based
visualization~\cite{van2008visualizing} of the poses in our training and testing sets.


\section{Conclusion}

We introduced a novel method for building high-quality photorealistic full-body
avatars that integrates, in its construction, the specific modality of driving signal
that is available during the model's use. 
Information deficiency in the driving signal is accounted for by using a latent space that 
is disentangled from the driving signal, enabling the generation of diverse plausible 
configurations that remain faithful to the information contained in the driving signal. 
We showcase the capabilities of the model by applying it in two example scenarios 
where the driving signal is information deficient, demonstrating improved generalization 
and fidelity compared with other approaches.

\begin{acks}
We would like to thank Carsten Stoll for providing body 
tracking code, 
Georgios Pavlakos for the face fitting pipeline, 
Sahana Vijai for providing assets for our body rig, 
Breannan Smith for high-quality hand tracking,
and Anuj Pahuja for the hand fitting pipeline
and help on VR demo. 
\end{acks}

\bibliographystyle{ACM-Reference-Format}
\bibliography{bibliography}

\end{document}